\documentclass[runningheads]{llncs}

\usepackage{eccv}

\usepackage{eccvabbrv}

\usepackage{graphicx}
\usepackage{booktabs}
\usepackage{wrapfig}
\usepackage{float}
\usepackage{subcaption}
\usepackage[final]{pdfpages}

\newcommand{\R}{{\mathbb{R}}}
\usepackage{color}

\usepackage{tikz}

\newcommand{\para}[1]{\vspace{0.5em} \noindent \textbf{#1} \hspace{0.5em}}
\newcommand{\paranospace}[1]{\noindent \textbf{#1} \hspace{0.5em}}

\definecolor{ddf_pos_colour}{rgb}{1.0, 0.0, 1.0}
\definecolor{ddf_dir_colour}{rgb}{0.5019, 0.0, 0.0}

\usepackage[accsupp]{axessibility}  

\usepackage{hyperref}

\usepackage{orcidlink}

\begin{document}

\title{The Sky's the Limit: Relightable Outdoor Scenes via a Sky-pixel Constrained Illumination Prior and Outside-In Visibility}

\titlerunning{The Sky's the Limit}

\author{James~A.~D.~Gardner\inst{1}\orcidlink{0000-0002-9492-3708} \and
Evgenii Kashin\inst{1}\orcidlink{0000-0001-5099-7361} \and
Bernhard Egger\inst{2}\orcidlink{0000-0002-4736-2397} \and
William~A.~P.~Smith\inst{1}\orcidlink{0000-0002-6047-0413}}

\authorrunning{J.~Gardner et al.}

\institute{Department of Computer Science, The University of York, York, YO10 5DD, UK \\
    \email{\{james.gardner,evgenii.kashin,william.smith\}@york.ac.uk} \and
    Cognitive Computer Vision Lab, Friedrich-Alexander-Universität, Erlangen-Nürnberg, Erlangen, Germany \\
    \email{bernhard.egger@fau.de}}

\maketitle
\begin{center}
    \centering
    \vspace{-0.4cm}
    \captionsetup{type=figure}
    \makebox[\textwidth]{%
        \begin{tikzpicture}
            \node (img) {\includegraphics[width=1.0\textwidth]{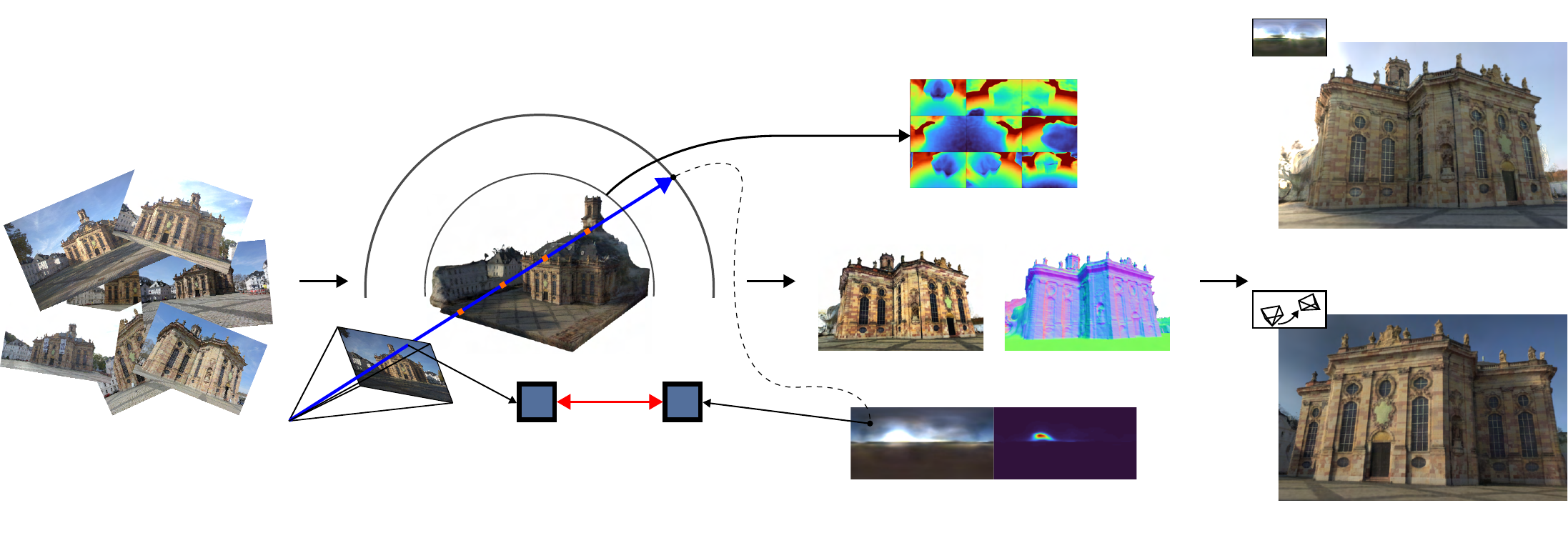}};
            \node at (-5.0,-1.3) {\tiny Unconstrained Outdoor};
            \node at (-5.0,-1.5) {\tiny Multi-View Images};
            \node at (-1.35 ,-1.25) {\scalebox{0.4}{Sky Pixel Illumination Constraint}};
            \node[align=center] at (-1.92,-1.7) {\tiny NeuS-Facto Volume};
            \node at (-1.92,0.89) {\scalebox{0.4}{Scene Geometry}};
            \node at (-1.92,0.69) {\scalebox{0.4}{Bounds}};
            \node at (-1.92,1.35) {\scalebox{0.4}{Sky at}};
            \node at (-1.92,1.15) {\scalebox{0.4}{Infinity}};
            \node at (0.9, -0.75) {\tiny Albedo};
            \node at (2.4, -0.75) {\tiny Normal};
            \node at (1.7, 0.5) {\tiny Differentiable Sky Visibility};
            \node[align=center] at (1.7,-1.75) {\tiny HDR Neural Illumination Prior};
            \node at (5.0, 0.2) {\tiny Relighting};
            \node at (5.0,-1.9) {\tiny Novel Views};
        \end{tikzpicture}
    }
    \vspace{-0.7cm}
    \captionof{figure}{From in-the-wild, outdoor image collections, we predict scene geometry, albedo, distant environment illumination, and sky visibility. Sky visibility and illumination are both modelled via spherical neural fields whereby we directly constrain illumination via sky pixel observations. Our outside-in differentiable visibility enables estimation of cast shadows and avoids shadow baking into albedo.}
    \label{fig:teaser}
    \vspace{-0.2cm}
\end{center}%

\begin{abstract}
Inverse rendering of outdoor scenes from unconstrained image collections is a challenging task, particularly illumination/albedo ambiguities and occlusion of the illumination environment (shadowing) caused by geometry. However, there are many cues in an image that can aid in the disentanglement of geometry, albedo and shadows.  Whilst sky is frequently masked out in state-of-the-art methods, we exploit the fact that any sky pixel provides a direct observation of distant lighting in the corresponding direction and, via a neural illumination prior, a statistical cue to derive the remaining illumination environment. The incorporation of our illumination prior is enabled by a novel `outside-in' method for computing differentiable sky visibility based on a neural directional distance function. This is highly efficient and can be trained in parallel with the neural scene representation, allowing gradients from appearance loss to flow from shadows to influence the estimation of illumination and geometry. Our method estimates high-quality albedo, geometry, illumination and sky visibility, achieving state-of-the-art results on the NeRF-OSR relighting benchmark.
Our code and models can be found at \url{https://github.com/JADGardner/neusky}.
\end{abstract}

\section{Introduction}
\label{sec:introduction}

Inverse rendering of outdoor scenes has diverse downstream applications such as scene relighting, augmented reality, game asset generation, and environment capture for films and virtual production. However, accurately estimating the underlying scene model that produced an image is an inherently ambiguous task due to its ill-posed nature \cite{barron_shape_2015}. To address this, many works use some combination of handcrafted \cite{bousseau2009user, grosse2009ground} or learned priors \cite{yu_inverserendernet_2019,bell2014intrinsic, boss2020two, kovacs2017shading}, inductive biases in model architectures \cite{davePANDORAPolarizationAidedNeural2022}, or multi-stage training pipelines \cite{srinivasan_nerv_2021, zhang_nerfactor_2021, sunSOLNeRFSunlightModeling2023}. This process is made even more difficult when considering in-the-wild image collections from the internet that contain transient objects, image filters, unknown camera parameters and changes in illumination. 

Outdoor scenes present particular challenges. Natural illumination from the sky is complex and exhibits an enormous dynamic range. This causes strong cast shadows when the brightest parts of the sky are occluded. These occlusions are non-local and discontinuous making them hard to incorporate within a differentiable renderer. Outdoor scene geometry can also exhibit arbitrary ranges of scale.
On the other hand, sky illumination dominates secondary bounce lighting, meaning it is reasonable to assume a spatially non-varying, distant illumination environment. In addition, natural illumination contains statistical regularities \cite{drorStatisticalCharacterizationRealworld2004} that make it easier to model. For example, luminance generally increases with elevation (the `lighting-from-above' prior), the sun can only be in one position and the range of possible colours from sun and sky light is limited.

In this paper, we tackle the outdoor scene inverse rendering problem by fitting a neural scene representation to a multi-view, varying-illumination photo collection. We name our method \textit{NeuSky}, and make four key contributions relative to prior work. First, we make a key observation: Any pixel in an image that observes the sky provides a direct constraint on the illumination environment in that direction. Second, we combine this insight with an HDR neural field natural illumination model \cite{gardner_reni++_2023} learnt from natural environments, constraining this model to outpaint plausible illuminations given the direct observations of illumination seen from the camera. Thirdly, we propose \textit{outside-in visibility}, a novel, differentiable, neural approximation to sky visibility, computed with a single forward pass through a directional distance function network. Finally, we deploy this visibility representation to enable end-to-end training, removing the need for phased training. Crucially, this means that shadows can influence illumination and geometry estimation by appearance losses backpropagating through the visibility network, enabling geometry estimation for non-observed scene regions and also avoiding shadow baking into albedo.

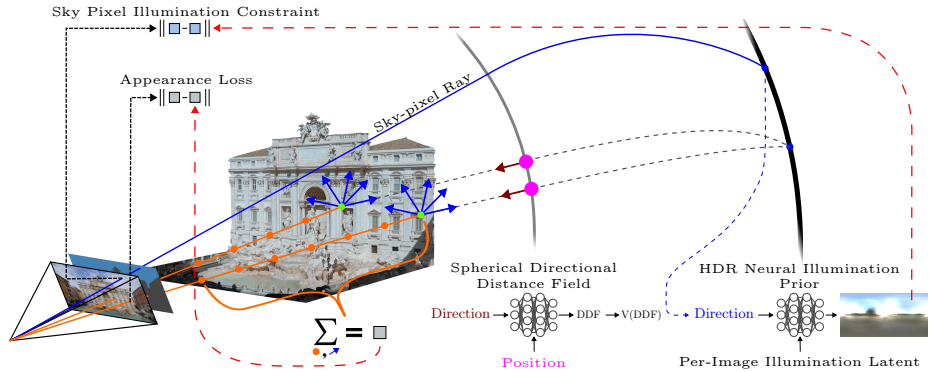
\begin{figure*}
    \centering
    \makebox[\textwidth]{%
        \begin{tikzpicture}
            \node (img) {\includegraphics[width=1.0\textwidth]{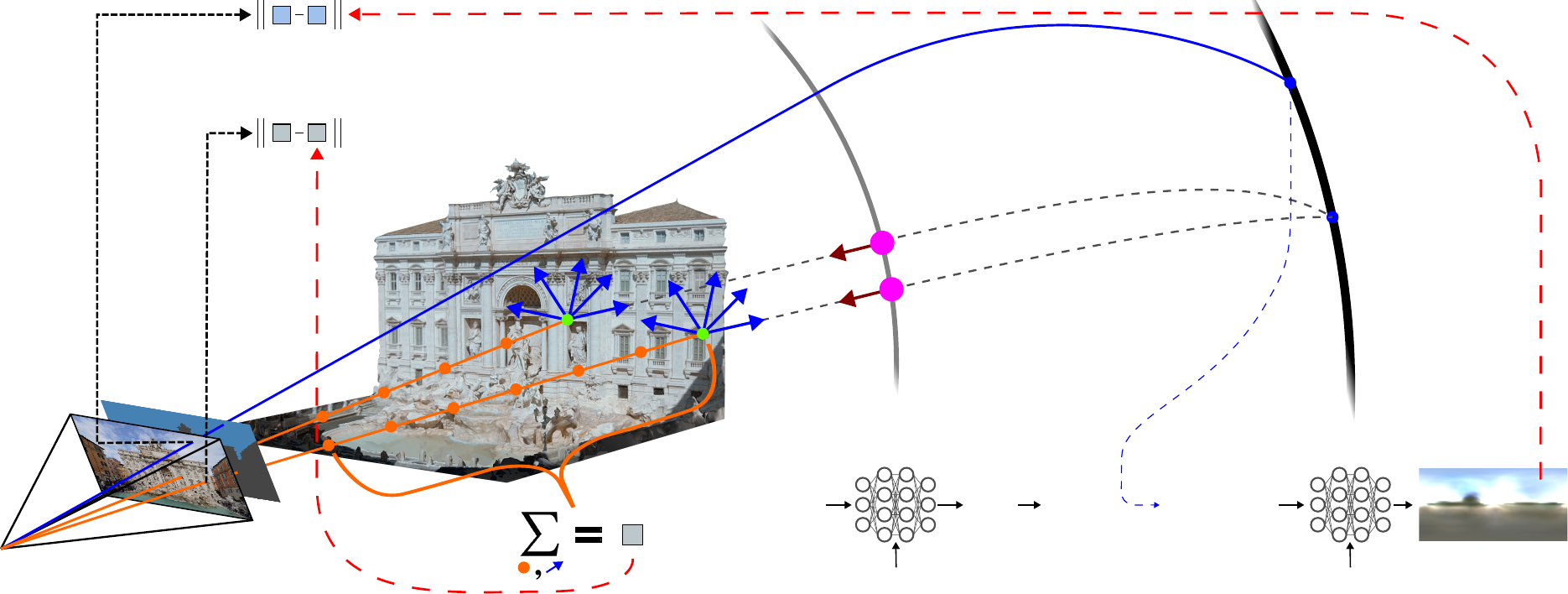}};
            \definecolor{ddf_pos_colour}{rgb}{1.0, 0.0, 1.0}
            \definecolor{ddf_dir_colour}{rgb}{0.5019, 0.0, 0.0}
            \node[text=ddf_pos_colour] at (0.87,-2.25) {\tiny Position};
            \node[text=ddf_dir_colour] at (-0.1,-1.6) {\scalebox{0.6}{Direction}};
            \node[text=blue] at (3.4,-1.6) {\scalebox{0.6}{Direction}};
            \node at (-3.75, 2.4) {\tiny Sky Pixel Illumination Constraint};
            \node at (-3.75, 1.5) {\tiny Appearance Loss};
            \node at (4.4, -1.0) {\tiny HDR Neural Illumination};
            \node at (4.4, -1.2) {\tiny Prior};
            \node at (4.4, -2.25) {\tiny Per-Image Illumination Latent};
            \node at (0.87, -1.0) {\tiny Spherical Directional};
            \node at (0.87, -1.2) {\tiny Distance Field};
            \node at (1.6, -1.61) {\scalebox{0.45}{DDF}};
            \node at (2.3, -1.62) {\scalebox{0.45}{V(DDF)}};
            \node at (-0.6, 1.2) [rotate=29.5] {\tiny Sky-pixel Ray};
        \end{tikzpicture}
    }
    \caption{We surround our NeuS-Facto \cite{yu_sdfstudio_2022} volume with two spherical neural fields at radius $1$ and radius $\infty$ modelling sky visibility and distant illumination respectively. \textcolor{blue}{Blue arrows} correspond to rays sampling distant illumination. \textcolor{ddf_pos_colour}{Pink circles} and \textcolor{ddf_dir_colour}{Maroon arrows} are position and direction samples of our sky visibility network. In a given direction, visibility changes with position but distant illumination does not. For speed we only sample sky visibility on the surface of our scene, \textcolor{green}{Green circles}, and distribute this visibility to all samples, \textcolor{orange}{Orange circles}, along a ray.} 
    \vspace{-3mm}
    \label{fig:technical_overview}
\end{figure*}

\section{Related Work}
\label{sec:related_work}

\paranospace{Relightable Neural Scenes} The core NeRF \cite{mildenhall_nerf_2020} approach has been improved in several key ways since its publication. Nerfstudio, \cite{tancik_nerfstudio_2023} a platform for researching in Neural Fields, introduced NeRFacto taking advantage of many of these developments. It leverages the same proposal sampling and scene contraction as Mip-NeRF 360 \cite{barron_mip-nerf_2022} alongside the hash-grid representation from Instant-NGP \cite{muller_instant_2022} to reduce network sizes and vastly speed up training. Implicit surface representations were introduced in NeuS \cite{wangNeuSLearningNeural2021} and VolSDF \cite{yarivVolumeRenderingNeural2021}, which used a neural Signed Distance Function (SDF) with NeRF volume rendering. NeuS-Facto, introduced in SDFStudio \cite{yu_sdfstudio_2022}, combined the NeRFacto improvements with NeuS. This model, which is similar to that used by the current state-of-the-art in neural surface reconstruction of large scenes \cite{liNeuralangeloHighFidelityNeural2023}, is the underlying model that we use.

In parallel with these developments, several attempts have been made to use neural scene representations for decomposition into its intrinsic properties. NeRF-OSR \cite{rudnev_nerf_2022} predicts albedo and density. For distant illumination, they predict per image Spherical Harmonic (SH) lighting coefficients and model shadows via a shadow network conditioned on those SH coefficients. Whilst now providing a parametric model of illumination they are limited by the quality of normals obtained from NeRF density (we use a NeuS derivative with high-quality geometry), shadows that are not related to the scene geometry (our shadow network is directly tied to scene geometry) and the low frequency of SH (we employ a neural field for illumination capable of capturing higher order lighting effects). Methods such as PhySG \cite{zhang2021physg} and NeRF-V \cite{srinivasan_nerv_2021} allow relighting but require known illumination. NeRFactor \cite{zhang_nerfactor_2021} additionally optimises visibility and illumination together allowing shadows but with a low-resolution environment map and no illumination prior. Similar to our work, FEGR \cite{wang_neural_2023} also uses a neural field representation for HDR illumination, however, they do not include a prior over illuminations. Their rasterisation process to model visibility is also a non-differentiable function, meaning cues from shading and shadows will not inform illumination or geometry estimations. SOL-NeRF \cite{sunSOLNeRFSunlightModeling2023} similarly convert their SDF representation to a mesh for ray-tracing but instead use a combination of Spherical Gaussians (SG), with a sunlight colour prior based on sun elevation, and SH to model illumination. Also similar to our work, NeuLighting \cite{liNeuLightingNeuralLighting2022} uses a prior over illuminations and visibility MLP but their framework is trained in a cascaded manner, so visibility can not influence lighting and geometry estimations compared to our method, furthermore their method considers shadows only from the sun. 

\para{Directional Distance Fields}
SDFs measure the distance to the nearest surface at a given point, signed to indicate outside/inside. In contrast, Directional Distance Functions (DDFs) measure the distance to the nearest surface \emph{in a given direction}, making them 5D as opposed to 3D functions for SDFs. Interest in DDFs has primarily been as a geometry representation that allows faster rendering (no sphere tracing is required). Neural DDFs were primarily introduced in \cite{zobeidiDeepSignedDirectional2021a} which developed the Signed Directional Distance Functions (SSDF) as a model of continuous distance view synthesis and derived many important properties of SDDFs. This was later extended by \cite{aumentado-armstrongRepresenting3DShapes2022}, which enabled the modelling of internal structures via dropping the sign and extending the representation via probabilistic modelling. Subsequent works enable to model of shapes with no explicit boundary surface \cite{uedaNeuralDensityDistanceFields2022}, refine the multi-view consistency of DDFs \cite{liu2023raydf} and employ SDDFs to improve optimisation of multi-view shape reconstruction \cite{zinsMultiViewReconstructionUsing2023}. Our usage of a DDF is most similar to that of FiRE \cite{yenamandraFIReFastInverse2022} which also combines an SDF scene representation with a DDF sampled only on the unit sphere. However, unlike FiRE, whose goal was fast rendering, we show how to use a spherical DDF for fast, differentiable sky visibility. A more in-depth explanation of DDFs is found in Section \ref{sec:ddf_visibility}.

\para{Neural Illumination and Visibility} Boss et al.~\cite{boss_neural-pil_2021} proposed neural pre-integrated lighting (PIL), a spherical neural field conditioned on a roughness parameter to model an illumination environment convolved with a BRDF. This enabled fast rendering but at the expense of being unable to model occlusions of the illumination environment. RENI ~\cite{gardner_rotation-equivariant_2022}, proposed by Gardner et al., is a vertical-axis rotation-equivariant conditional spherical neural field, trained on thousands of HDR outdoor environment maps to learn a prior for natural illumination. The low-dimensional but expressive latent space is useful for constraining inverse rendering problems. This was subsequently extended in RENI++ \cite{gardner_reni++_2023} with the addition of scale-invariant training and a transformer-based architecture. Several other recent methods aim to predict illumination from small image crops \cite{Somanath_2021_CVPR, dastjerdiEverLightIndoorOutdoorEditable2023}, as a 5D light field network \cite{yaoNeILFNeuralIncident2022}, from a text description \cite{chenText2LightZeroShotTextDriven2022} or using diffusion models with differentiable path tracing \cite{lyu2023dpi}. Rhodin et al.~\cite{Rhodin:2015} approximate scene geometry with Gaussian blobs for differentiable visibility. Lyu et al.~\cite{lyu2021efficient} similarly use spheres for geometry and model illumination with spherical harmonics for approximate differentiable shadows. Worchel and Alexa \cite{worchel2023differentiable} use a differentiable mesh renderer for classical shadow mapping \cite{williams1978casting}. 

\section{Method}
\label{sec:method}

Our method takes as input a dataset of $N$ images. From these images we compute poses with COLMAP \cite{schoenberger2016sfm, schoenberger2016mvs} and semantic segmentation maps with ViT-Adapter \cite{chen_vision_2023} according to the Cityscapes \cite{cordts_cityscapes_2016} convention. The preprocessed dataset comprises $\mathcal{D} = \left\{(\mathcal{I}_{i}, \mathbf{E}_{i}, \mathbf{K}_{i}, \mathcal{S}_{i}) \right\}_{i=1}^{N}$, where $\mathcal{I} \in \R^{H \times W \times 3}$ is an image, $\mathcal{S} \in \mathbb{Z}^{H \times W}$ is the segmentation map and $\mathbf{E} = \left [ \mathbf{R} | \mathbf{t} \right ] \in \R^{3 \times 4}$ and $\mathbf{K} \in \R^{3 \times 3}$ are the camera extrinsics and intrinsics respectively. To align the vertical axis of our scene with gravity, we robustly fit a plane to the camera positions and rotate to align with the $x$-$y$ plane.

\para{Scene Representation} We model scene geometry as a neural SDF, such that at any point $\mathbf{x} \in \R^{3}$, the signed distance is given by $f_\text{SDF}(\mathbf{x})\in\R$. We assume that the scene is Lambertian, with diffuse albedo modelled by the neural field $\mathbf{a}(\mathbf{x})\in[0,1]^3$. We also assume that illumination is a distant environment that depends only on direction $\mathbf{d}\in S^2$, with HDR RGB incident radiance given by $L_i(\mathbf{d})\in\R_{\geq 0}^3$.

\para{Rendering} We follow NeuS \cite{wang_neus_2021} and derive a volume density, $\sigma(\mathbf{x})$, from the SDF value. This allows volume rendering of the SDF in the same fashion as in NeRF. For a ray $\mathbf{r}$ with origin $\mathbf{o}$ and direction $\mathbf{v}$, the time-discrete volume rendered RGB colour is given by:
\begin{equation}
    \mathbf{c}(\mathbf{r})\!=\!\! \sum_{j=1}^S \!w_j\mathbf{a}(\mathbf{x}_j) \!\sum_{k=1}^D  \!L_i(\mathbf{d}_k) V(\mathbf{x}_E,\mathbf{d}_k)\! \max(0,\mathbf{n}(\mathbf{x}_j)\cdot\mathbf{d}_k),\label{eqn:rendering}
\end{equation}
where the first summation is over the $S$ samples along the ray, while the second is over the $D$ lighting direction samples. The lighting direction samples are distributed approximately uniformly over the sphere by using an $8$-subdivided icosahedron giving $D=642$. $w_j$ is the volume rendering blending weight for the $j$th sample point which depends on $t_{1\dots j}$ and $\sigma(\mathbf{x}_{1\dots j})$, with $\mathbf{x}_j=\mathbf{o}+t_j\mathbf{v}$. $V(\mathbf{x},\mathbf{d})\in\{0,1\}$ is the sky visibility in direction $\mathbf{d}$ at position $\mathbf{x}$ with $\mathbf{x}_E$ being the position at the expected termination depth of the ray $\mathbf{x}_j$. $\mathbf{n}(\mathbf{x})=\nabla f_\text{SDF}(\textbf{x}) / \|\nabla f_\text{SDF}(\textbf{x})\|$ is the surface normal at $\mathbf{x}$, derived from the gradient of the SDF.

We define our appearance loss for a batch of rays $\mathcal{R}$ as:
\begin{equation}
    \mathcal{L}_\text{app}=\sum_{\mathbf{r}\in\mathcal{R}} \ell(\mathbf{c}_\text{gt}(\mathbf{r}),\text{sRGB}(\mathbf{c}(\mathbf{r}))),
\end{equation}
where $\mathbf{c}_\text{gt}(\mathbf{r})$ is the ground truth colour for ray $\mathbf{r}$, $\text{sRGB}(\cdot)$ tonemaps the linear image provided by our model and $\ell$ computes the sum of L1 and cosine errors (to match both absolute RGB values and hue). To avoid overfitting we apply a random rotation $R \sim \mathcal{U}(SO(3))$ to jitter the direction vectors $\mathbf{d}_k$ in every batch.

\para{Neural Illumination Model} To restrict $L_i$ to the space of plausible natural illumination environments, we use a neural illumination prior, RENI++ \cite{gardner_reni++_2023}. This is a conditional neural field, $f_{L_i}:S^2\times\R^{3\times K}\rightarrow \R^3$ that outputs log HDR RGB colours in the given input direction, conditioned on a normally distributed 3D latent code $\mathbf{Z}\in\R^{3\times K}, \text{vec}(\mathbf{Z})\sim\mathcal{N}(\mathbf{0}_{3K},\mathbf{I}_{3K})$. The latent space of RENI++ provides a low dimensional characterisation of natural, outdoor illumination environments since it was trained on several thousand real-world outdoor environment maps. This provides useful global constraints on the estimated illumination, which is only partially observable in any one image. In addition, the normally-distributed latent space provides a prior while the latent code is vertical-axis rotation-equivariant (rotating $\mathbf{Z}$ about the vertical axis corresponds to similarly rotating the environment). This vertical axis corresponds to gravity and we therefore align the vertical axis of our scene with gravity as described above. We optimise a RENI++ latent code, $\mathbf{Z}_i$, and absolute scale, $\gamma_i$, for each image $i$ in the training set and replace $L_i(\mathbf{d}_k)$ with $\gamma_i \exp(f_{L_i}(\mathbf{d}_k,\mathbf{Z}_i))$ in \eqref{eqn:rendering}. To ensure the estimated illumination is plausible, we include a prior loss: $\mathcal{L}_\text{prior}=\|\mathbf{Z}\|_2^2$ for all latent codes. In Section \ref{sec:sky_pixel} we describe how the illumination environment in an image can be additionally constrained via sky pixel observations.

\para{Reducing Visibility Tests} Visibility of the illumination environment from a scene point is required in our rendering equation \eqref{eqn:rendering} and is essential for recreating cast shadows and ambient occlusion effects. However, computing sky visibility from a neural SDF is computationally expensive. It requires sphere tracing from the query point in the light direction until the ray hits another part of the surface or leaves the scene bounds. To render a single pixel, this must be performed $D$ times for each of the $S$ sample points. We therefore propose two methods to drastically reduce the number of visibility samples. First, since we are only concerned with visibility on the surface of the scene, we define $\mathbf{x}_E=\mathbf{o}+t_E\mathbf{v}$, where $t_E$ is the current expected termination depth of the ray, and evaluate visibility only at $\mathbf{x}_E$. This means we only need $D$ visibility tests per pixel since we reuse the computed visibilities for all sample points along the ray. Second, any light direction in the lower hemisphere, i.e.~where $(\mathbf{d}_j)_z<0$, will strike either the scene or the ground. For these directions we set $V(\cdot)=1$, i.e.~visible. The rationale for this is that the RENI++ illumination environment will learn to capture the colour of the ground or lower hemisphere of the scene, averaged over all spatial positions. This provides an approximation to secondary illumination from the ground. We found this to perform considerably better than setting these directions as non-visible. In spite of these two speedups, the remaining $D/2$ visibility tests still prove too expensive if performed via sphere tracing of the SDF. For this reason, in Section \ref{sec:ddf_visibility} we propose a fast, softened approximation for visibility.

\subsection{Sky Pixel Constrained Illumination Prior}
\label{sec:sky_pixel}

Pixels labelled in the semantic segmentation maps with the `sky' class (hereafter referred to as \emph{sky pixels}) provide a direct observation of the distant illumination environment in the direction given by the ray for that pixel. To the best of our knowledge, this constraint has never been used to aid illumination estimation in inverse rendering methods. Since our illumination model, RENI++ \cite{gardner_reni++_2023}, captures the space of plausible natural illuminations, even observing only a portion of the sky provides a strong statistical cue. For example, if a bright region corresponding to the sun is observed, then RENI++ cannot create another sun in an unobserved part of the environment. Alternatively, if all observed sky is white, it is likely to be an overcast day and RENI++ will predict an ambient environment without a discernible sun. Using sky pixel constraints alone can be viewed as statistical outpainting of the whole environment from the portion observed in an image. In practice, we incorporate this within our inverse rendering framework such that the appearance loss of non-sky pixels also provides a rich, indirect constraint on the illumination.

The sky segmentation also provides an additional constraint that is similar to the widely used mask loss. Since we know that sky ray pixels miss the scene, we penalise our neural scene representation from placing any density along the ray, providing geometric supervision. Together, these form our sky loss:
\begin{equation}
    \mathcal{L}_\text{sky} = \!\!\!\sum_{\mathbf{r}\in\mathcal{R}\cap\mathcal{S}_\text{sky}}\!\!\! \varepsilon(\mathbf{c}_\text{gt}(\mathbf{r}),\mathbf{c}_\text{sky}(\mathbf{r})) - 
    \log( 1-\sum_{j} w_j ),
\end{equation}
where $\mathcal{S}_\text{sky}$ is the set of sky pixels. The first term is the error between the observed sky pixel colour and predicted, $\mathbf{c}_\text{sky}(\mathbf{r})=\text{sRGB}(\gamma\exp(f_{L_i}(\mathbf{r},\mathbf{Z})))$, and the second term is the binary cross entropy loss on the accumulated density in sky pixels.
\subsection{Outside-in Sky Visibility}
\label{sec:ddf_visibility}

Shadows offer a wealth of information about geometry, both within and beyond the view frustum. For instance, if the sun is predicted to be behind the camera and a prominent cast shadow appears on the floor, we can infer there is geometry behind the camera and the likely sun direction. However, to fully leverage this information it is necessary to have a differentiable model of visibility.

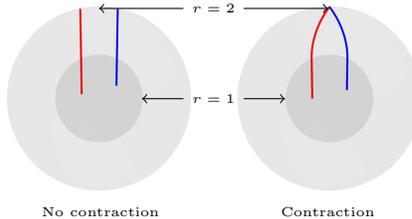
\begin{wrapfigure}{r}{0.5\textwidth}
  \centering
  \vspace{-8mm}
  \makebox[0.5\textwidth]{%
    \begin{tikzpicture}
        \node (img) {\includegraphics[width=0.5\textwidth]{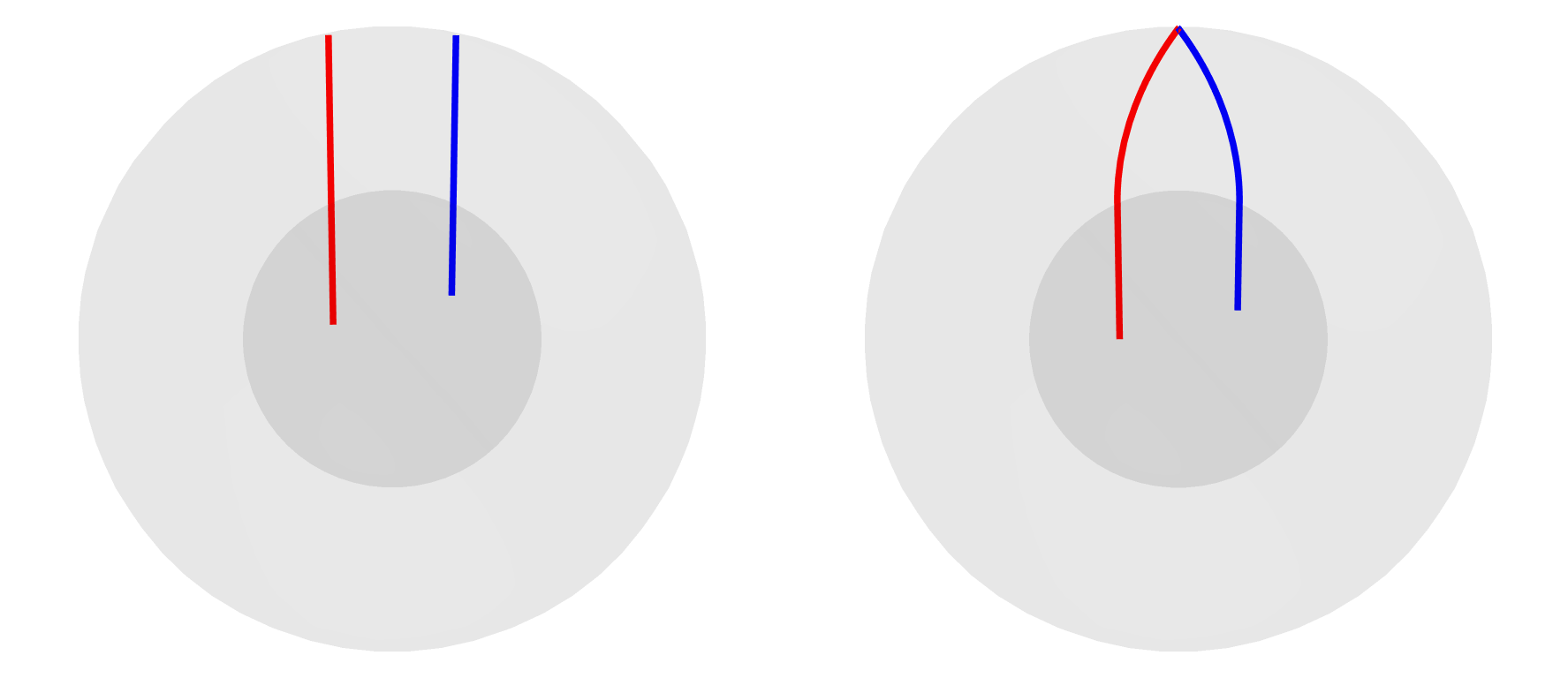}};
        \draw[->] (0.4,1.2) -- (1.52,1.2);
        \draw[->] (-0.4,1.2) -- (-1.52,1.2);
        \node at (0.0, 1.2) {\tiny $r = 2$};
        \node at (0.0, 0.0) {\tiny $r = 1$};
        \draw[->] (0.4,0.00) -- (0.95,0.00);
        \draw[->] (-0.4,0.00) -- (-0.95,0.00);
        \node at (-1.5, -1.5) {\tiny No contraction};
        \node at (1.52, -1.5) {\tiny Contraction};
    \end{tikzpicture}
  }
    \caption{We model our illumination and illumination visibility via two spherical neural fields at radius $\infty$ and $1$ respectively. However our world space is contracted as per Mip-NeRF-360 \cite{barron_mip-nerf_2022}, such that any point at infinity is placed on the sphere of radius $2$. Since we model distant illumination, the sampled colour only depends on direction, and two samples at different locations but in the same direction will sample RENI++ \cite{gardner_reni++_2023} at the same point. However, visibility of distant illumination \textit{is} dependent on location and the intersection of the ray on the sphere of radius $1$ is used to sample our visibility network.}
    \vspace{-6mm}
  \label{fig:contraction}
\end{wrapfigure}

To address this, we draw inspiration from works, NeRFactor \cite{zhang_nerfactor_2021} and NeRV \cite{srinivasan_nerv_2021} and learn a neural model of visibility. However, to make training tractable, \cite{zhang_nerfactor_2021} learn their visibility representation in a second training phase with geometry pretrained and frozen and \cite{srinivasan_nerv_2021} require known illumination. Initial attempts to model visibility using the same parameterisation as \cite{srinivasan_nerv_2021} were unable to fit in our less constrained and end-to-end task. We desire a model of visibility that is consistent with the geometry of our scene, fast to sample from and differentiable, enabling gradients from visibility to inform illumination, albedo and geometry estimation. However, this model must be constrained enough that training end-to-end with our scene representation is tractable. To achieve this, we propose \emph{outside-in visibility} in which visibility is represented implicitly via a Spherical Directional Distance Field (SDDF) defined on the radius 1 sphere that bounds our scene and is tied to our SDF scene representation via consistency losses. Our geometric volume is represented with the Mip-NeRF 360 \cite{barron_mip-nerf_2022} scene contraction. This means that parallel rays converge to a point on the radius 2 sphere (representing infinity). Hence, our visibility model resides on the radius 1 sphere where position-dependent visibility can be reasoned about, while our distant illumination model is defined on the radius 2 sphere (see Figure \ref{fig:contraction}).

\para{Spherical Directional Distance Function}
Consider a point $\mathbf{s}\in S^2$ lying on a bounding sphere of radius~$1$. The Spherical Directional Distance Function (DDF),
$f_\text{DDF} : S^{2} \times S^{2} \to \R$,
returns the (positive) distance from $\mathbf{s}$ for any inward-pointing direction $\mathbf{d}$ to the first intersection with the surface. In other words, the spherical DDF stores an inward looking depth map of the scene from any viewpoint on the radius $r$ sphere. The DDF is related to the SDF:
$f_\text{SDF}(\mathbf{s}+f_\text{DDF}(\mathbf{s},\mathbf{d})\mathbf{d}) = 0$,
such that moving the distance given by the DDF must arrive at the surface where the SDF is zero. However, there may be multiple such points and the DDF must return the minimum, giving us another constraint:
$f_\text{DDF}(\mathbf{s},\mathbf{d}) = \min \{ t | f_\text{SDF}(\mathbf{s}+t\mathbf{d})=0 \}$.

\begin{figure}
    \centering
    \makebox[\textwidth]{%
        \begin{tikzpicture}
            \node (img) {\includegraphics[width=0.7\textwidth]{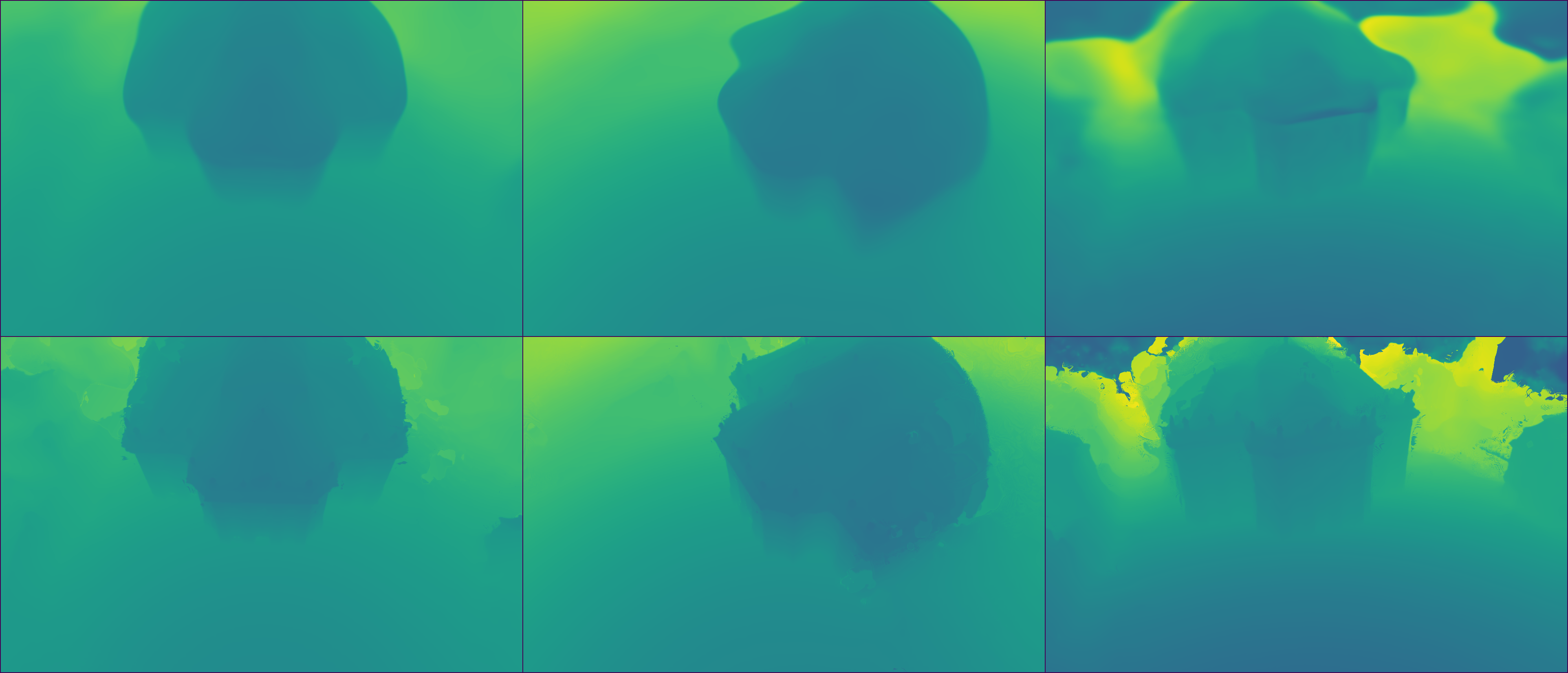}};
            \node at (-4.45, 0.9) [rotate=90] {\tiny DDF Prediction};
            \node at (-4.45, -0.9) [rotate=90] {\tiny SDF Target};
        \end{tikzpicture}
    }
    \caption{Three views of the depth predicted by the Spherical DDF (top row) and its pseudo ground truth from the scene representation (bottom row) for \textit{Site 1} in the NeRF-OSR \cite{rudnev_nerf_2022} dataset (see Figure \ref{fig:teaser}). Cameras are placed on the unit sphere looking towards the origin. The DDF is trained concurrently with the scene representation and can capture high-frequency details required for accurate shadows.}
    \label{fig:ddf_depth}
\end{figure}

The DDF is required to learn a very complex function: essentially an inward-facing depth map of the scene from any position on the sphere. We found that this function is easier to learn if we define a consistent coordinate frame to parameterise directions for any given point on the sphere. We normalise the inward-facing directions from world coordinates to a local coordinate system such that the $y$-axis aligns with $\mathbf{s}$ (the sample position on the DDF), the $x\text{-axis}$ is orthogonal to $y$ and to our world-up, and the $z$-axis is orthogonal to $y$ and $x$. See Figure \ref{fig:ddf_depth} for a visualisation.

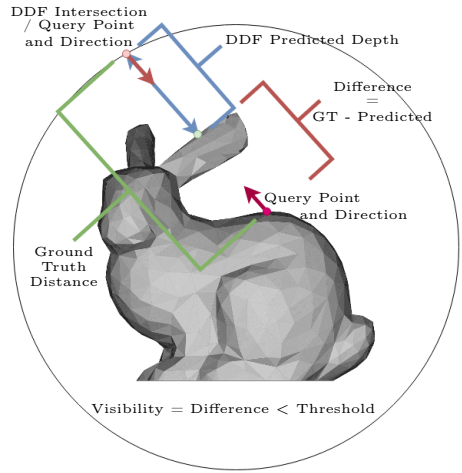
\begin{wrapfigure}{r}{0.5\textwidth}
  \centering
  \vspace{-5mm}
  \makebox[0.5\textwidth]{%
        \begin{tikzpicture}
                \node (img) {\includegraphics[width=0.5\textwidth]{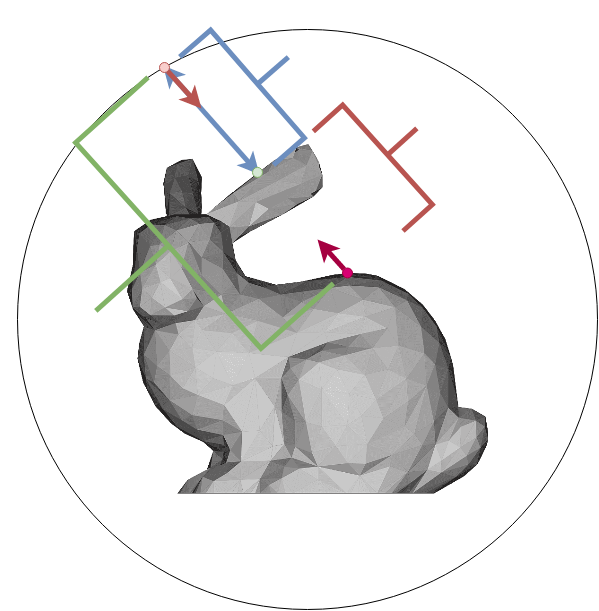}};
                \node at (-2.0, 3.0) {\tiny DDF Intersection};
                \node at (-2.0, 2.8) {\tiny / Query Point};
                \node at (-2.0, 2.6) {\tiny and Direction};
                \node at (1.1, 2.6) {\tiny DDF Predicted Depth};
                \node at (1.1, 0.5) {\tiny Query Point};
                \node at (1.65, 0.3) {\tiny and Direction};
                \node at (-2.2, -0.2) {\tiny Ground};
                \node at (-2.2, -0.4) {\tiny Truth};
                \node at (-2.2, -0.6) {\tiny Distance};
                \node at (1.9, 2.0) {\tiny Difference};
                \node at (1.9, 1.8) {\tiny =};
                \node at (1.9, 1.6) {\tiny GT - Predicted};
                \node at (0.05, -2.3) {\tiny Visibility = Difference $<$ Threshold};
        \end{tikzpicture}
    }
    \caption{Visibility of our neural illumination from a point in the scene is implicitly represented via our Directional Distance Field (DDF) which represents the depth to the surface of our scene from any point on the unit sphere. The DDF is a spherical neural field that surrounds our scene at radius~$1$. The DDF is fully differentiable allowing gradients obtained from shadowing to inform illumination and geometry.}
    \vspace{-6mm}
    \label{fig:implicit_visibility}
\end{wrapfigure}

\para{Sky Visibility via Directional Distance Fields} 
Our key insight is to show how to use the inward looking DDF as a representation for computing outward sky visibility (see Figure \ref{fig:implicit_visibility}).
Consider a point $\mathbf{x}\in\R^3$ lying on the surface (and inside the bounding sphere, such that $\|\mathbf{x}\|\leq 1$). We can use the DDF to check whether $\mathbf{x}$ can see the sky or is occluded in a direction $\mathbf{d}$. First we compute the point $\mathbf{s}$ as the solution to $\mathbf{s}=\mathbf{x}+t\mathbf{d}$, s.t.~$\|\mathbf{s}\|=1$ and $t\geq 0$, i.e.~the point on the radius $r$ sphere that is intersected by the ray in direction $\mathbf{d}$ from $\mathbf{x}$. Next, we evaluate the DDF at $\mathbf{s}$ in direction $-\mathbf{d}$ (i.e.~outside-in): $f_\text{DDF}(\mathbf{s},-\mathbf{d})$. If $\mathbf{x}$ is not occluded then the DDF value should be similar to the actual distance between $\mathbf{s}$ and $\mathbf{x}$: $f_\text{DDF}(\mathbf{s},-\mathbf{d})\approx \|\mathbf{s}-\mathbf{x}\|$. However, if $\mathbf{s}$ is occluded then the DDF will return a distance significantly less than the actual distance: $f_\text{DDF}(\mathbf{s},-\mathbf{d}) < \|\mathbf{s}-\mathbf{x}\|$. Binary visibility can be computed by testing whether this difference is below a threshold $\epsilon$: $V=(\|\mathbf{s}-\mathbf{x}\|-f_\text{DDF}(\mathbf{s},-\mathbf{d})<\epsilon)$. Note that this is equivalent to classical shadow mapping \cite{williams1978casting} with the exception that we rely on a DDF forward pass as opposed to (non-differentiable) rasterisation of a mesh from the light source perspective. 

\begin{wrapfigure}{r}{0.53\textwidth}
    \centering
    \vspace{-4mm}
    \includegraphics[width=0.53\textwidth]{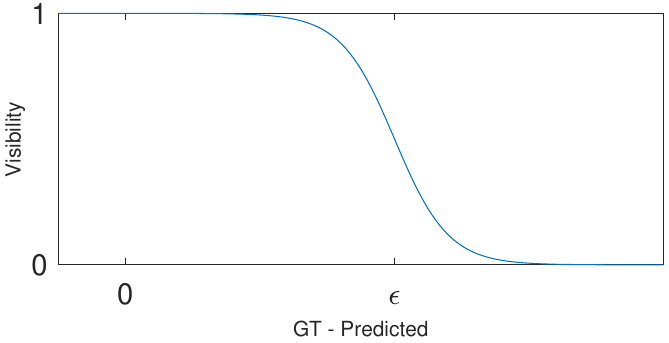}
    \caption{Soft visibility function. We plot $\|\mathbf{s}-\mathbf{x}\|-f_\text{DDF}(\mathbf{s},-\mathbf{d})$ on the $x$-axis versus $V(\mathbf{x},\mathbf{d})$ on the $y$-axis. When ground truth distance is significantly smaller than the threshold $\epsilon$, we assign a visibility of 1. When significantly larger, we infer an occlusion and a visibility of 0. In the vicinity of $\epsilon$ we smoothly transition from visible to non-visible with a steepness controlled by $\eta$.}
    \vspace{-5mm}
    \label{fig:visibilitysigmoid}
\end{wrapfigure}

However, binary visibility is discontinuous and so not suitable for propagating loss gradients through visibility and back into geometry. For this reason, we replace the discrete threshold with a softened approximation (see Figure \ref{fig:visibilitysigmoid}):
\begin{equation}
    V(\mathbf{x},\mathbf{d}) = 1-\kappa\left( \eta(\|\mathbf{s}-\mathbf{x}\|-f_\text{DDF}(\mathbf{s},-\mathbf{d})-\epsilon) \right),
\end{equation}
where $\kappa$ is the sigmoid function. The threshold $\epsilon$ controls the tolerance on what is considered a shadow. We make this learnable and initialise it with a large value (equal to the scene radius). When $\epsilon$ is large, no parts of the scene will be considered occluded. As training converges, $\epsilon$ can be reduced to gradually introduce more illumination occlusions. The parameter $s$ controls the sharpness of the transition between occluded and unoccluded.

\para{Supervising the DDF} 
The DDF indirectly determines visibility which in turn determines appearance via the rendering equation in \eqref{eqn:rendering}. This means that the DDF is partially supervised by the appearance loss. However, we also require that the DDF's representation of scene geometry is consistent with the SDF geometry. We enforce this consistency through four losses. First, $\mathcal{L}_\text{ddf\_depth}$, enforces that the depth predicted by the DDF should match that of the scene parameterised by the SDF. Second, $\mathcal{L}_\text{ddf\_levelset}$, ensures that travelling the distance predicted by the DDF should arrive at the SDF zero level set. Third, we encourage multiview consistency in the DDF via a multiview consistency loss $\mathcal{L}_\text{ddf\_multiview}$. Finally, with, $\mathcal{L}_\text{ddf\_sky}$, we further take advantage of our sky segmentation maps as an additional constraint on our DDF. Rays that intersect the sky have no occlusions between the camera origin and our DDF sphere. Our DDF should therefore predict at least the distance to the camera origin for those intersecting rays. Detailed descriptions of these losses can be found in the supplementary.
\newpage
\subsection{Implementation}
\label{sec:implementation}

We implement our method in Nerfstudio \cite{tancik_nerfstudio_2023}, building on top of NeuS-Facto \cite{yu_sdfstudio_2022}. We convert our CityScapes \cite{cordts_cityscapes_2016} segmentation masks into classes for sky, ground plane, foreground and transient objects (vehicles, vegetation, people etc). We sample ray batches only from non-transient pixels. We use a hash grid with $16$ levels, $2^{19}$ hash table size, $2$ features per entry and a course and fine resolution of $16$ and $2048$ respectively. Our SDF and albedo networks are both $2$-layer $256$-neuron MLPs. We initialise our SDF as a sphere with radius=$0.1$. We use the pre-trained RENI++ \cite{gardner_reni++_2023} model with a latent dimension $K = 100$ and initialise latent codes as zeroes, corresponding to the mean environment provided by the RENI++ prior. We initialise the per-image illumination scale as $\gamma=1$.

Our visibility network is a FiLM-Conditioned \cite{chanPiGANPeriodicImplicit2021} SIREN \cite{sitzmannImplicitNeuralRepresentations2020} with $5$ layers and $256$ neurons in both the FiLM Mapping Network and the main SIREN. We condition our network on positions on the sphere using the same dimensional hash grid as our SDF. We first map from position to a hashed latent, this latent is then provided to the FiLM mapping network to condition the model. As per Section \ref{sec:ddf_visibility} we normalise direction to a local coordinate frame and these directions are positionally encoded as per NeRF \cite{mildenhall_nerf_2020}. We use a sigmoid activation function scaled by the size of our scene bounds to ensure a depth prediction within the correct range. We generate samples for DDF supervision via our PyTorch re-implementation of fast von Mises-Fisher distribution sampling from Pinzón et al \cite{pinzonFastPythonSampler2023a}. We used a concentration parameter of $20.0$ for the distribution and sampled $8$ positions and $128$ directions per batch exclusively from the upper hemisphere.

We optimise our Proposal Samplers, SDF/Albedo Field and DDF using Adam \cite{kingmaAdamMethodStochastic2015} optimisers with a Cosine Decay \cite{loshchilov2017sgdr} schedule and 500-step `warm-up' phase. Our loss is the sum of $\mathcal{L}_\text{app}$, $\mathcal{L}_\text{prior}$, $\mathcal{L}_\text{sky}$, the four DDF supervision losses and a proposal sampler interlevel loss as per Mip-NeRF 360 \cite{barron_mip-nerf_2022}. Our initial learning rates are $1\text{e}{-2}$, $1\text{e}{-3}$ and $1\text{e}{-4}$ respectively. Our RENI++ latent codes and the visibility threshold parameter use Adam \cite{kingmaAdamMethodStochastic2015} optimisers with an exponentially decaying learning rate which is initialised at $1\text{e}{-2}$ and $1\text{e}{-3}$ respectively.
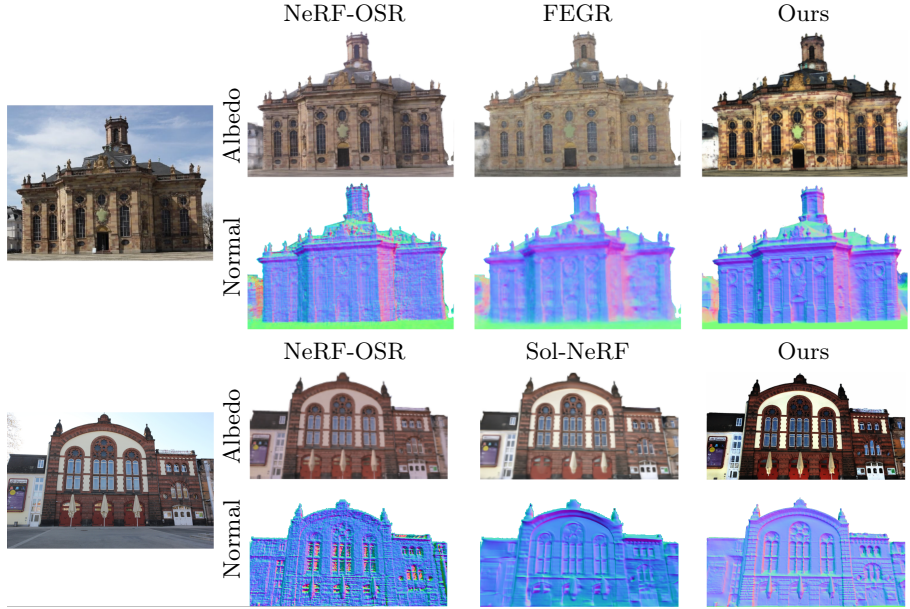
\begin{figure*}
    \centering
    \makebox[\textwidth]{%
        \begin{tikzpicture}
            \node (img) {\includegraphics[width=0.98\textwidth]{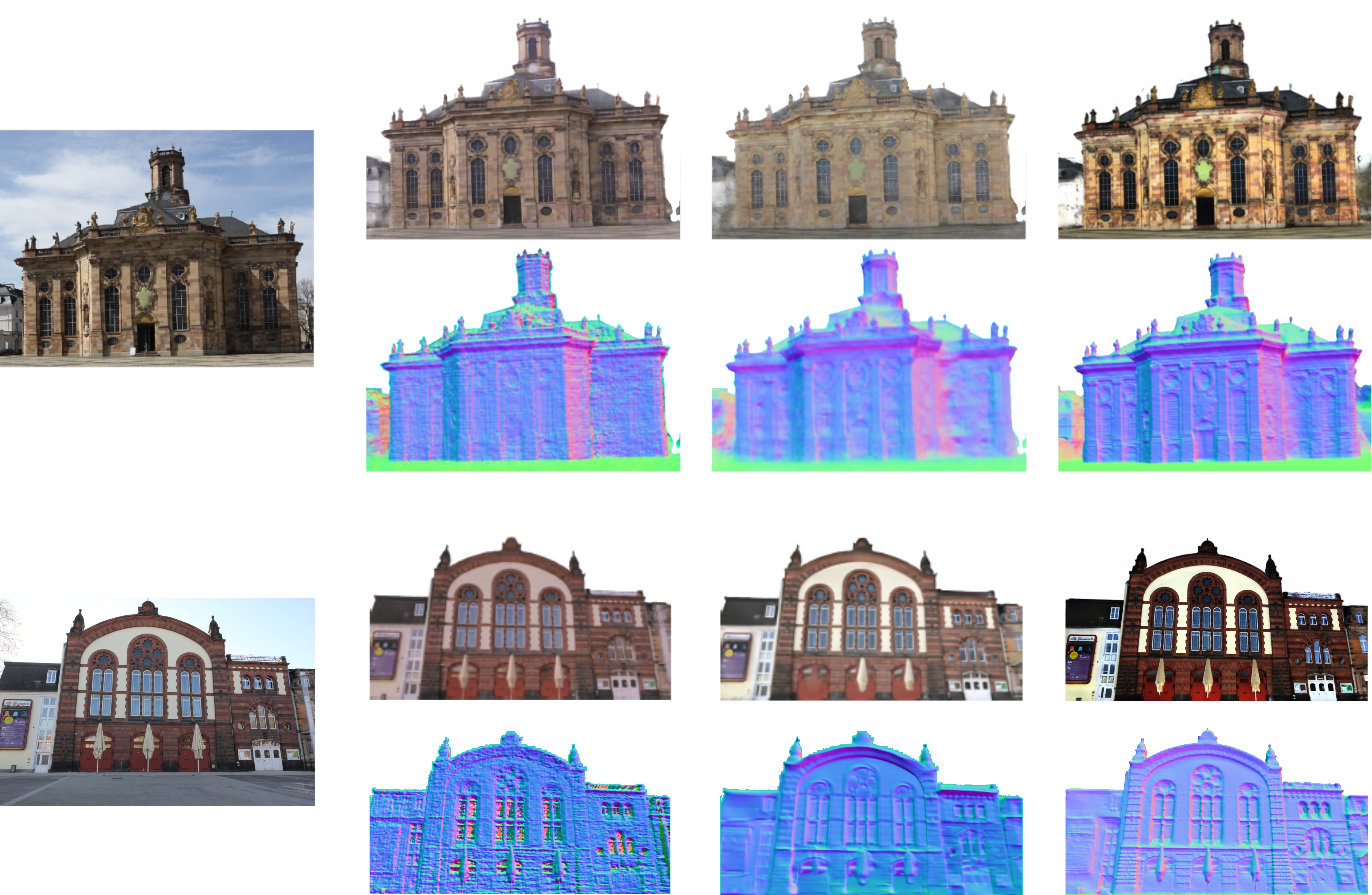}};
            \node at (-1.5, 4.0) {\small NeRF-OSR};
            \node at (1.6, 4.0) {\small FEGR};
            \node at (4.6, 4.0) {\small Ours};
            \node at (-1.5, -0.5) {\small NeRF-OSR};
            \node at (1.6, -0.5) {\small Sol-NeRF};
            \node at (4.6, -0.5) {\small Ours};
            \node at (-3.0, 2.5) [rotate=90] {\small Albedo};
            \node at (-3.0, 0.75) [rotate=90] {\small Normal};
            \node at (-3.0, -1.5) [rotate=90] {\small Albedo};
            \node at (-3.0, -3.0) [rotate=90] {\small Normal};
        \end{tikzpicture}
    }
    \caption{Comparion of albedo and normals produced by NeRF-OSR \cite{rudnev_nerf_2022}, FEGR \cite{wang_neural_2023}, SOL-NeRF \cite{sunSOLNeRFSunlightModeling2023} and our method. We produce much sharper albedo and normals than all prior works whilst training end-to-end.}
    \label{fig:comparison}
\end{figure*}

\newpage
\section{Evaluation}
\label{sec:evaluation}

\begin{wrapfigure}{r}{0.5\textwidth}
  \centering
  \vspace{-5mm}
  \makebox[0.5\textwidth]{%
    \begin{tikzpicture}
        \node (img) {\includegraphics[width=0.5\textwidth]{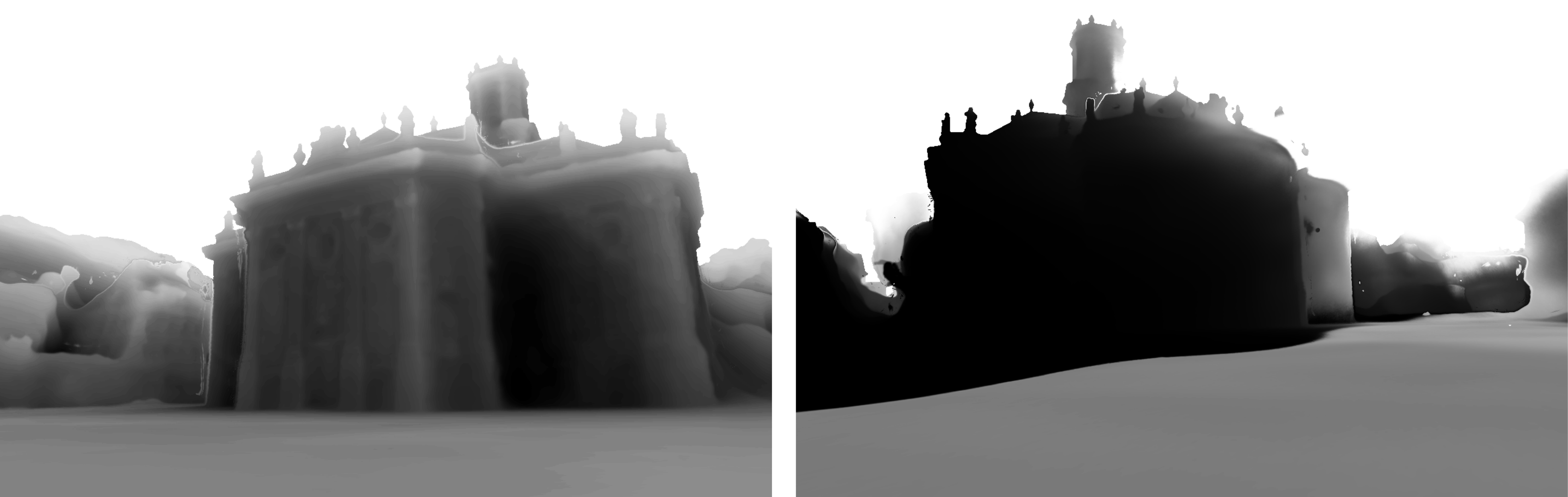}};
        \node at (-1.6, 1.1) {\tiny Ambient Occlusion};
        \node at (1.6, 1.1) {\tiny Shadow};
    \end{tikzpicture}
  }
    \caption{Ambient occlusion and shadows from a point source computed from our soft visibility via the DDF.}
    \vspace{-6mm}
  \label{fig:ao_and_shadow}
\end{wrapfigure}

We begin by qualitatively evaluating the output of our system components. Figure \ref{fig:ddf_depth} illustrates that our spherical DDF is able to produce detailed depth maps via a single forward pass from arbitrary viewpoints. The geometry of the building and ground plane are well reconstructed. In Figure \ref{fig:ao_and_shadow} we visualise the output of the visibility network in two different ways. On the left we average visibility over all directions, giving a good approximation to ambient occlusion. On the right we compute visibility for a single direction, producing a sharp shadow.

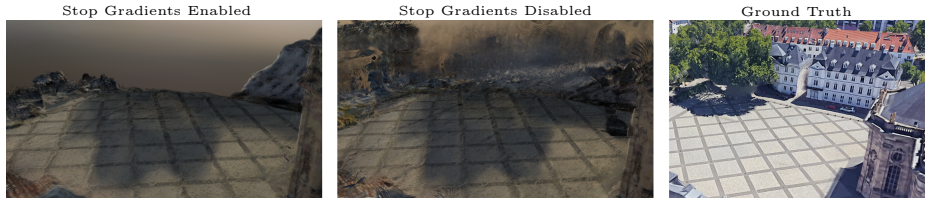
\begin{figure*}
    \centering
    \makebox[\textwidth]{%
    \begin{tikzpicture}
        \node (img) {\includegraphics[width=\textwidth]{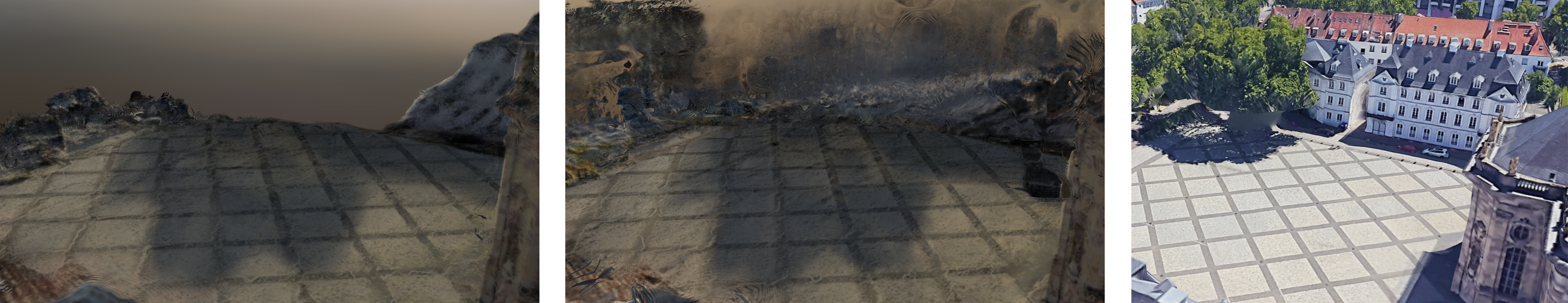}};
        \node at (-4.1, 1.3) {\tiny Stop Gradients Enabled};
        \node at (0.4, 1.3) {\tiny Stop Gradients Disabled};
        \node at (4.4, 1.3) {\tiny Ground Truth};
    \end{tikzpicture}
    }
    \caption{With stop gradients enabled (left), geometry outside the view frustum of all training cameras is not generated. With stop gradients disabled (middle), gradients from appearance losses are allowed to flow from our sky visibility network to the SDF creating geometry not directly observed during training. This more closely matches the ground truth for the scene (right).}
    \label{fig:geometry_outside_cameras}
\end{figure*}

\para{Shadows Informing Geometry} Due to our visibility model training concurrently with our scene representation, shadows can inform geometry outside of the view frustum. We can optionally apply stop gradients to visibility calculations to prevent this capability. We demonstrate the advantage of training end-to-end in Figure~\ref{fig:geometry_outside_cameras}, which shows a rendering of \textit{Site 1} looking behind the view frustums of all training cameras for that scene. To explain shadows seen during training geometry has been generated outside the view of any training camera. This is a key advantage of training our differentiable sky visibility network concurrently with our scene representation.

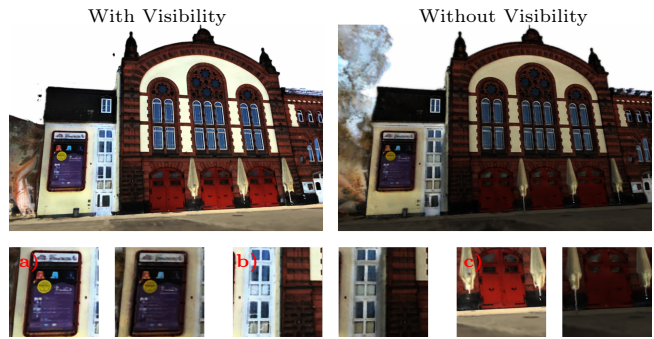
\begin{figure*}
    \centering
    \makebox[\textwidth]{%
    \begin{tikzpicture}
        \node (img) {\includegraphics[width=0.7\textwidth]{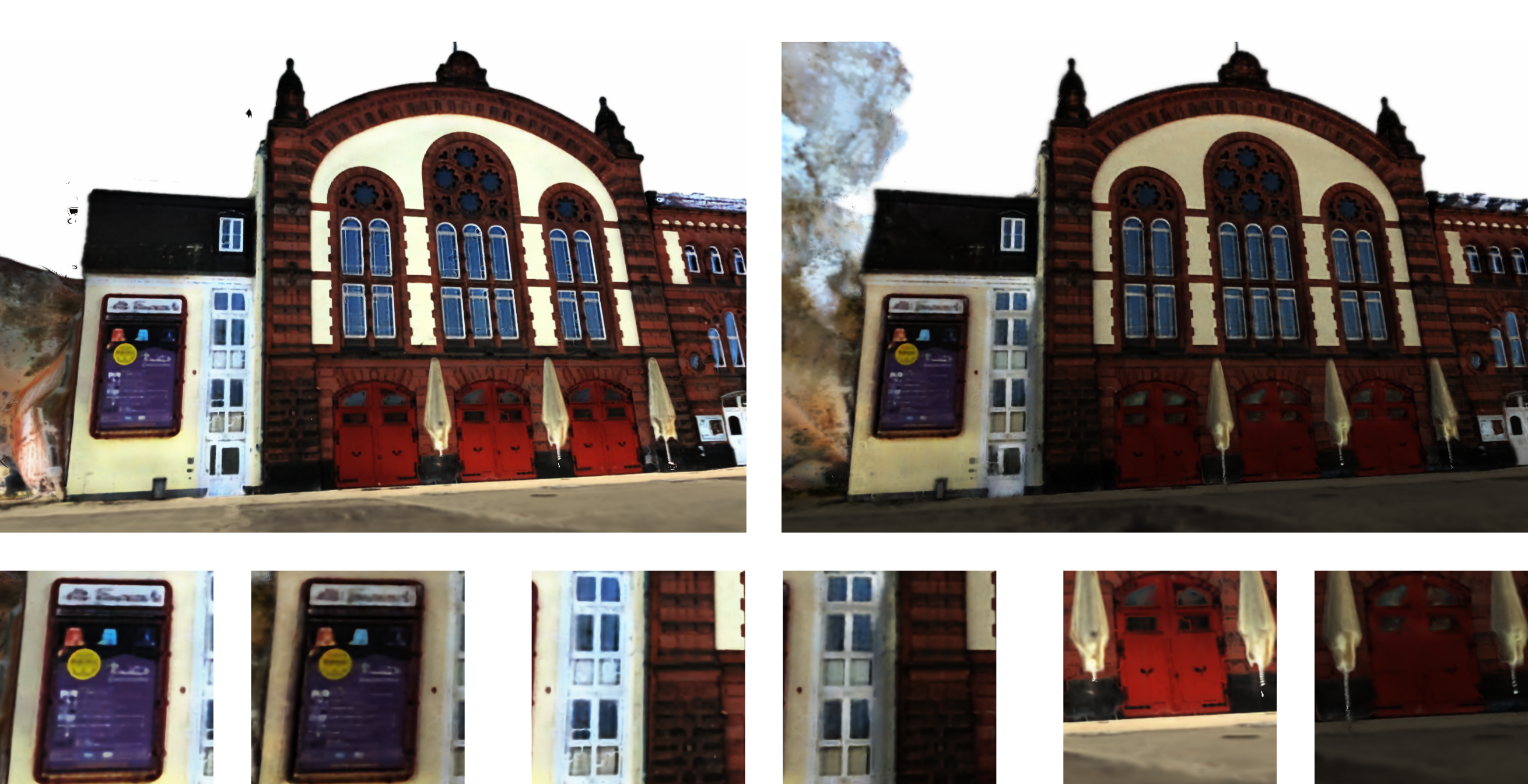}};
        \node at (-2.3, 2.05) {\scriptsize With Visibility};
        \node at (2.3, 2.05) {\scriptsize Without Visibility};
        \node[text=red] at (-4.0,-1.2) {\scriptsize \textbf{a)}};
        \node[text=red] at (-1.1,-1.2) {\scriptsize \textbf{b)}};
        \node[text=red] at (1.9,-1.2) {\scriptsize \textbf{c)}};
    \end{tikzpicture}
    }
    \caption{Whilst the majority of the training images for \textit{Site 3} in the NeRF-OSR dataset \cite{rudnev_nerf_2022} show the front of the building in shadow. With our visibility network enabled our predicted albedo removes that shading along with shadows around the sign \textit{(a)}, at the joint between brick and plaster \textit{(b)} and on the ground \textit{(c)}. Smaller cutouts show renderings with sky visibility on the left and without sky visibility on the right.}
\label{fig:albedo_comparison}
\end{figure*}

\para{Without Visibility Network} Figure \ref{fig:albedo_comparison} demonstrates the benefit of our sky-visibility network. When enabled our model is better able to disentangle shading from albedo, particularly in scenes in which many of the images captured are shaded, namely \textit{Site 3} of the NeRF-OSR \cite{rudnev_nerf_2022} dataset. Here, shadows on the ground, doors and building facade are removed from the albedo. 

\begin{wrapfigure}{r}{0.5\textwidth}
\vspace{-6mm}
\centering 
\setlength{\tabcolsep}{4pt}
\renewcommand{\arraystretch}{1.2}
\resizebox{0.5\textwidth}{!}{ 
\begin{tabular}{@{}lcccccccc@{}}
\toprule
& \multicolumn{2}{c}{Site 1} & \phantom{ab} & \multicolumn{2}{c}{Site 2} & \phantom{ab} & \multicolumn{2}{c}{Site 3} \\
& PSNR $\uparrow$ & MSE $\downarrow$ && PSNR $\uparrow$ & MSE $\downarrow$ && PSNR $\uparrow$ & MSE $\downarrow$ \\
\midrule
NeRF-OSR~\cite{rudnev_nerf_2022} & 19.34 & 0.012 && 16.35 & 0.027 && 15.66 & 0.029 \\
FEGR~\cite{wang_neural_2023} & 21.53 & 0.007 && 17.00 & 0.023 && 17.57 & 0.018 \\
SOL-NeRF~\cite{sunSOLNeRFSunlightModeling2023} & 21.23 & 0.0084 && \textbf{18.18} & \textbf{0.019} && 17.58 & 0.028 \\
NeuSky (Ours) & \textbf{22.50} & \textbf{0.005} && 16.66 & 0.023 && \textbf{18.31} & \textbf{0.016} \\
\bottomrule
\end{tabular}
}
\caption{Outdoor scene relighting results on the \textit{NeRF-OSR} relighting benchmark.}
\vspace{-5mm}
\label{tab:nerf_osr}
\end{wrapfigure}

\begin{figure*}
    \vspace{-6mm}
    \centering
    \makebox[\textwidth]{%
        \begin{tikzpicture}
            \node (img) {\includegraphics[width=0.98\textwidth]{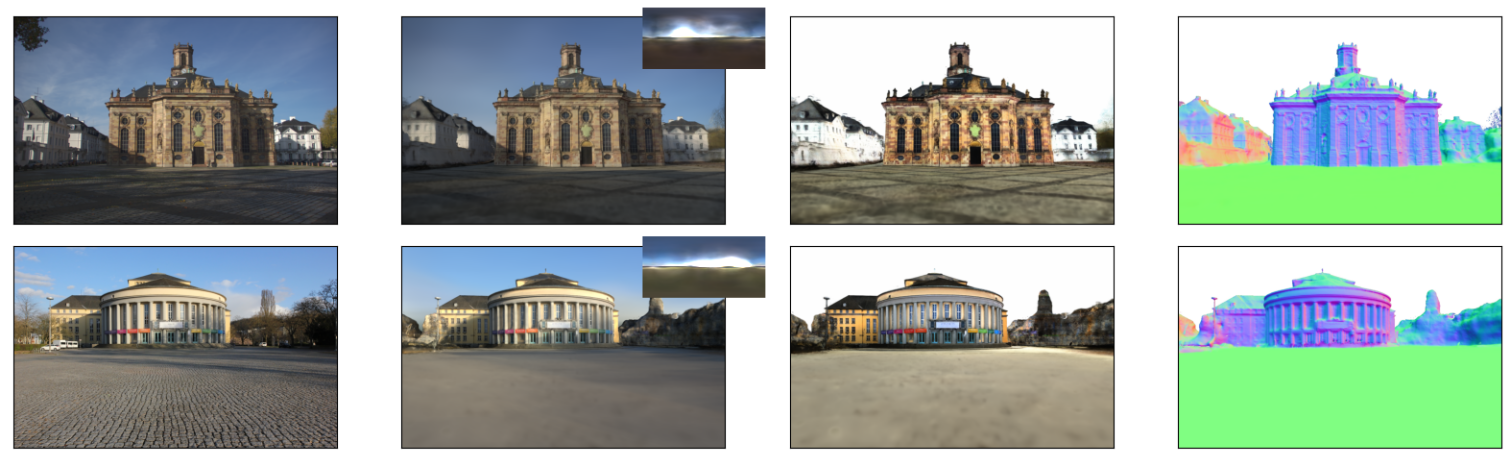}};
            \node at (-4.6, 2.0) {\small Ground Truth};
            \node at (-1.5, 2.0) {\small Render};
            \node at (1.5, 2.0) {\small Albedo};
            \node at (4.6, 2.0) {\small Normals};
        \end{tikzpicture}
    }
    \caption{A render from \textit{Site-1} and \textit{Site-2} in NeRF-OSR \cite{rudnev_nerf_2022}. Environment maps sampled from the estimated illumination of RENI++\cite{gardner_reni++_2023}, albedo and normals are shown alongside the ground truth images. Our method accurately disentangles albedo, lighting and shadows whilst producing very high-quality geometry.}
    \vspace{-5mm}
    \label{fig:examples}
\end{figure*}
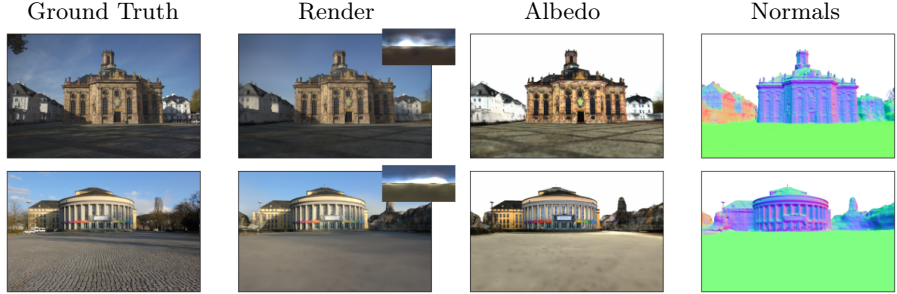

\begin{figure*}
    \vspace{-8mm}
    \centering
    \makebox[\textwidth]{%
        \begin{tikzpicture}
            \node (img) {\includegraphics[width=0.98\textwidth]{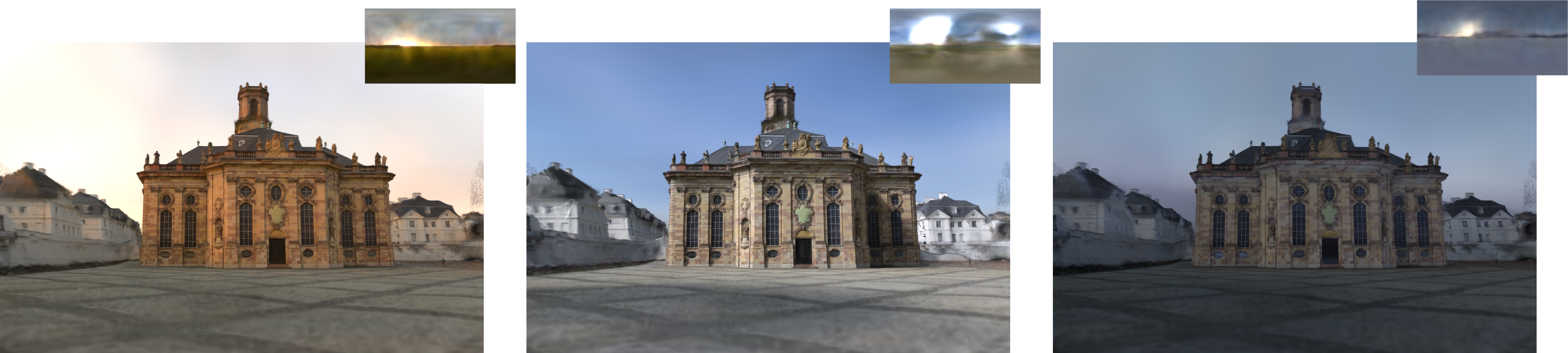}};
        \end{tikzpicture}
    }
    \caption{Relighting under novel illuminations.}
    \vspace{-12mm}
    \label{fig:relighting_examples}
\end{figure*}

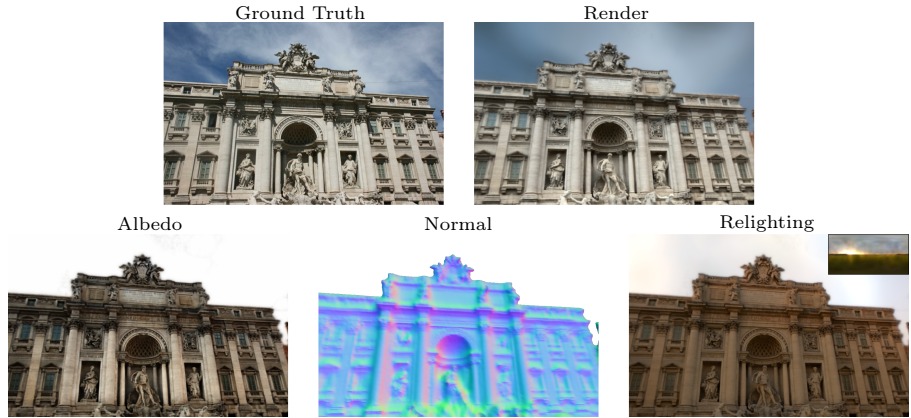
\begin{figure*}
    \centering
    \makebox[\textwidth]{%
        \begin{tikzpicture}
            \node (img) {\includegraphics[width=0.98\textwidth]{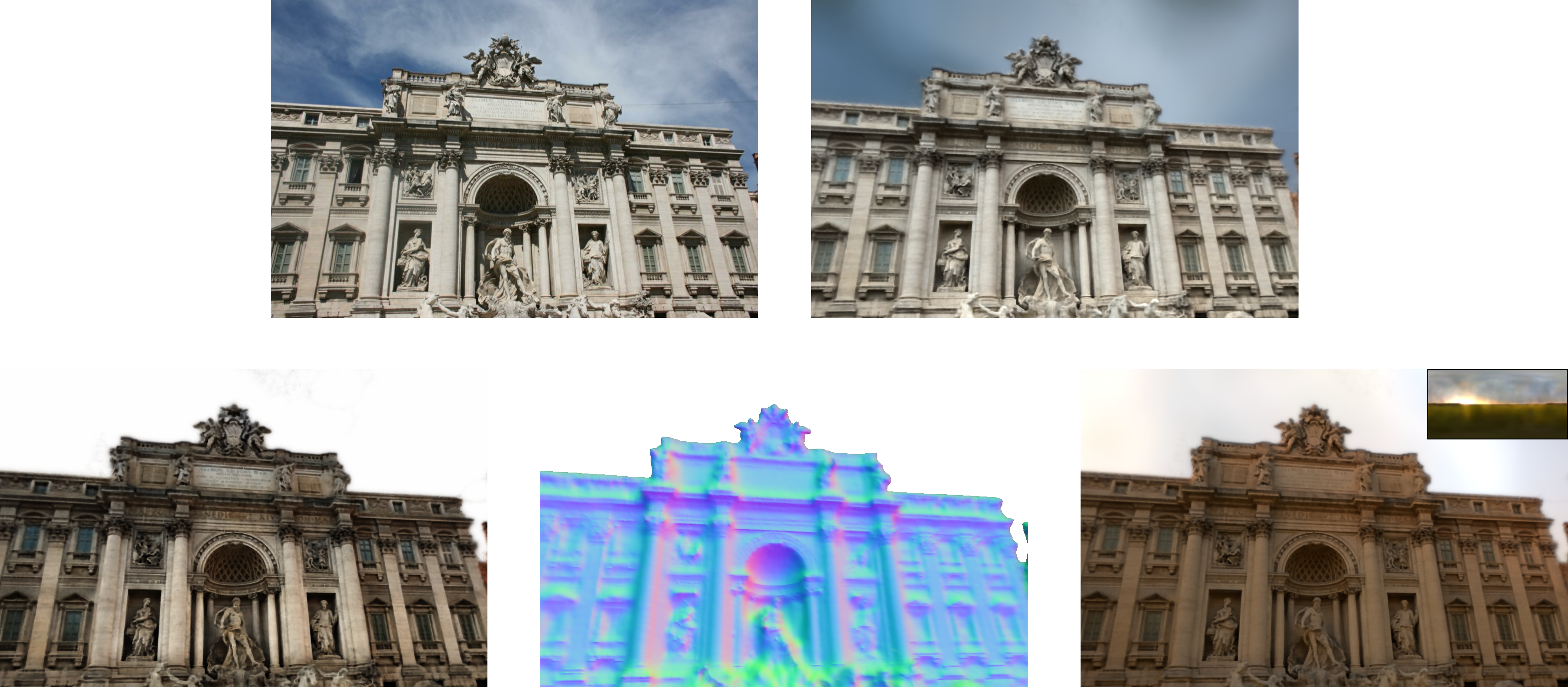}};
            \node at (-2.1, 2.75) {\scriptsize Ground Truth};
            \node at (2.1, 2.75) {\scriptsize Render};
            \node at (0.0, -0.05) {\scriptsize Normal};
            \node at (-4.1, -0.05) {\scriptsize Albedo};
            \node at (4.1, -0.05) {\scriptsize Relighting};
        \end{tikzpicture}
    }
    \vspace{-5mm}
    \caption{Results on the Trevi Fountain scene showing decomposition and relighting.}
    \label{fig:trevi}
\end{figure*}

\para{Relighting} We evaluate NeuSky's relighting capabilities on the NeRF-OSR \cite{rudnev_nerf_2022} relighting benchmark. The NeRF-OSR dataset consists of eight sites captured over multiple sessions each with differing illumination conditions along with Low Dynamic Range (LDR) ground truth environment maps. The benchmark test relighting three of these scenes. 
Our results can be found in Table \ref{tab:nerf_osr}. We achieve better relighting performance than both NeRF-OSR \cite{rudnev_nerf_2022} and FEGR \cite{wang_neural_2023} and beat SOL-NeRF \cite{sunSOLNeRFSunlightModeling2023} on two of the three scenes. Qualitatively our method also produces models with significantly higher quality geometry and albedo than all three prior works, as shown in Figure \ref{fig:comparison}. As shown in Figures \ref{fig:examples} and \ref{fig:trevi}, NeuSky is capable of disentangling illumination, albedo and shading and our sky visibility network and RENI++ combine to produce sharp shadows. Further results are shown in Figure \ref{fig:teaser}. In Figure \ref{fig:relighting_examples} we demonstrate rendering scenes under novel illumination conditions.

\section{Conclusion}
\label{sec:conclusion}

We have presented the first outdoor scene inverse rendering approach that incorporates a model of natural illumination, exploits direct sky pixel observations and can be trained end-to-end with a visibility model. This enables our model to reproduce accurate shadows, avoids shadow baking into albedo, allows shadows to constrain geometry and illumination and achieves superior geometry and albedo reconstruction on the NeRF-OSR dataset beating \cite{rudnev_nerf_2022}, \cite{sunSOLNeRFSunlightModeling2023} and \cite{wang_neural_2023} in the relighting benchmark. There are a number of limitations of our work, namely a high training GPU memory requirement when using large batches and at between 5-8 hours, our optimisation time is slow by modern neural field standards. The most obvious extension to our approach would be to use a more complex reflectance model and accompanying material parameters. For reflective surfaces, second bounce illumination becomes more significant. It is possible that a DDF could be used to speed up multibounce ray casting in this context.

\paragraph{Acknowledgments and Disclosure of Funding}
\label{s:Acknowledgments}

This project was supported by travel funds from the Bavarian Research Alliance (BayIntAn  FAU -2023-29). James Gardner and Evgenii Kashin were supported by the EPSRC CDT in Intelligent Games \& Games Intelligence (IGGI) (EP/S022325/1).

%
%
\bibliographystyle{splncs04}
\bibliography{references}

\includepdf[pages=-]{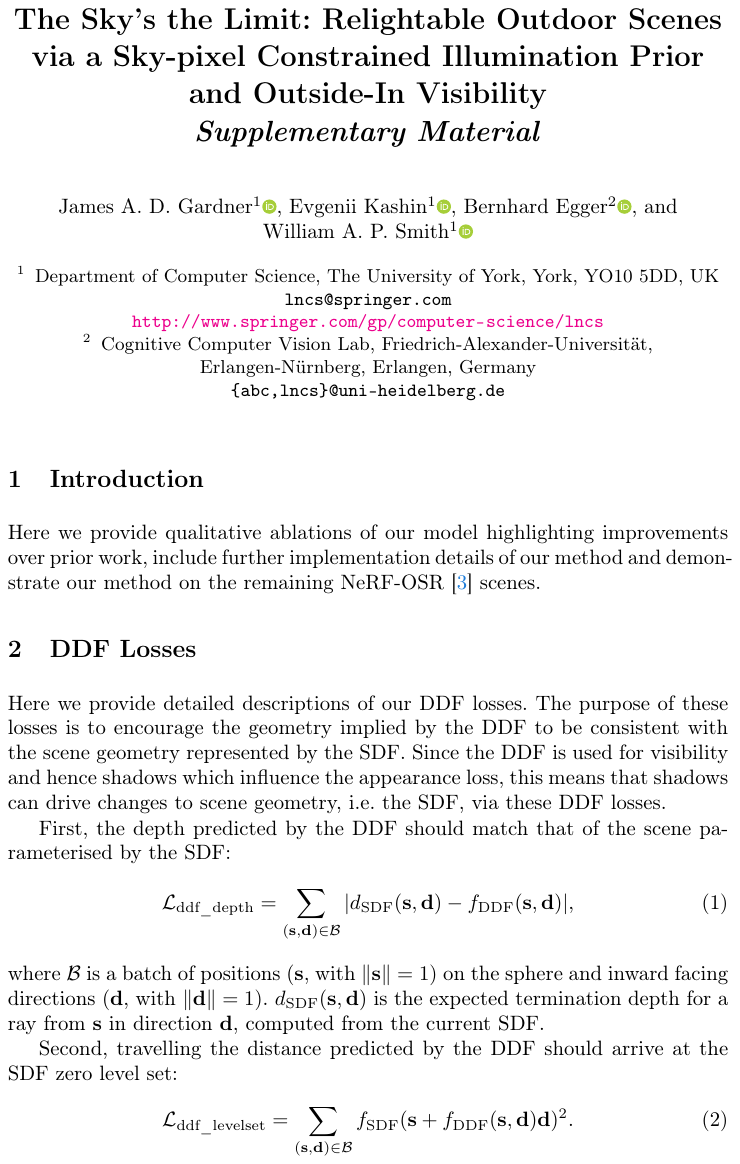}

\end{document}


\title{The Sky's the Limit: Relightable Outdoor Scenes via a Sky-pixel Constrained Illumination Prior and Outside-In Visibility\newline\textit{Supplementary Material}}

\titlerunning{The Sky's the Limit}

\author{James~A.~D.~Gardner\inst{1}\orcidlink{0000-0002-9492-3708} \and
Evgenii Kashin\inst{1}\orcidlink{0000-0001-5099-7361} \and
Bernhard Egger\inst{2}\orcidlink{0000-0002-4736-2397} \and
William~A.~P.~Smith\inst{1}\orcidlink{0000-0002-6047-0413}}

\authorrunning{J.A.D.~Gardner et al.}

\institute{Department of Computer Science, The University of York, York, YO10 5DD, UK
\email{lncs@springer.com}\\
\url{http://www.springer.com/gp/computer-science/lncs} \and
Cognitive Computer Vision Lab, Friedrich-Alexander-Universität, Erlangen-Nürnberg,  Erlangen, Germany\\
\email{\{abc,lncs\}@uni-heidelberg.de}}

\maketitle

\section{Introduction}
Here we provide qualitative ablations of our model highlighting improvements over prior work, include further implementation details of our method and demonstrate our method on the remaining NeRF-OSR \cite{rudnev_nerf_2022} scenes.

\section{DDF Losses}\label{sec:ddf_losses}
Here we provide detailed descriptions of our DDF losses. The purpose of these losses is to encourage the geometry implied by the DDF to be consistent with the scene geometry represented by the SDF. Since the DDF is used for visibility and hence shadows which influence the appearance loss, this means that shadows can drive changes to scene geometry, i.e.~the SDF, via these DDF losses.

First, the depth predicted by the DDF should match that of the scene parameterised by the SDF:
\begin{equation}
    \mathcal{L}_\text{ddf\_depth} = \sum_{(\mathbf{s},\mathbf{d})\in\mathcal{B}}  |d_\text{SDF}(\mathbf{s},\mathbf{d}) - f_\text{DDF}(\mathbf{s},\mathbf{d})|,
\end{equation}
where $\mathcal{B}$ is a batch of positions ($\mathbf{s}$, with $\|\mathbf{s}\|=1$) on the sphere and inward facing directions ($\mathbf{d}$, with $\|\mathbf{d}\|=1$). $d_\text{SDF}(\mathbf{s},\mathbf{d})$ is the expected termination depth for a ray from $\mathbf{s}$ in direction $\mathbf{d}$, computed from the current SDF.

Second, travelling the distance predicted by the DDF should arrive at the SDF zero level set:
\begin{equation}
\mathcal{L}_\text{ddf\_levelset} = \sum_{(\mathbf{s},\mathbf{d})\in\mathcal{B}} f_\text{SDF}(\mathbf{s} + f_{\text{DDF}}(\mathbf{s}, \mathbf{d})\mathbf{d})^2.
\end{equation}
This loss penalises any non-zero SDF value at the termination point predicted by the DDF.

Third, we can impose multiview consistency on the DDF. Given an arbitrary starting point $\mathbf{s}_1$ and inward facing direction $\mathbf{d}_1$, we compute a termination point $\mathbf{x}_1=\mathbf{s}_1 + f_{\text{DDF}}(\mathbf{s}_1, \mathbf{d}_1)\mathbf{d}_1$. Now, from an arbitrary second point $\mathbf{s}_2$, the predicted DDF depth towards $\mathbf{x}_1$ must be no greater than $\|\mathbf{x}_1-\mathbf{s}_2\|$, since $\mathbf{x}_1$ would occlude $\mathbf{s}_2$:
\begin{equation}
    \mathcal{L}_\text{ddf\_multiview} = \sum_{(\mathbf{s}_1,\mathbf{d}_1,\mathbf{s}_2)\in\mathcal{B}} \max\left(0,f_\text{DDF}(\mathbf{s}_2, \mathbf{d}_2)-\|\mathbf{x}_1-\mathbf{s}_2\|\right)^2,
\end{equation}
where $\mathbf{d}_2=(\mathbf{x}_1-\mathbf{s}_2)/\|\mathbf{x}_1-\mathbf{s}_2\|$ and this time the batch comprises pairs of points and a direction.

Finally, we further take advantage of our sky segmentation maps as an additional constraint on our DDF. Rays that intersect the sky have no occlusions between the camera origin and our DDF sphere. Our DDF should therefore predict at least the distance to the camera origin for those intersecting rays:
\begin{equation}
    \mathcal{L}_\text{ddf\_sky} = \sum_{\mathbf{r}\in\mathcal{R}\cap\mathcal{S}_\text{sky}} \max\left(0, \|\mathbf{o}-\mathbf{s}\|-f_\text{DDF}(\mathbf{s}, -\mathbf{r}) \right)
\end{equation}
where $\mathbf{s}$ is the point where the camera ray $\mathbf{r}$ intersects the DDF sphere and $\mathbf{o}$ the camera origin. Note that this last loss provides direct, ground truth supervision for the DDF as opposed to the previous three losses that only ensure consistency with the SDF. It plays the same role for the DDF as $\mathcal{L}_\text{sky}$ plays for the SDF, except it is used in an inward facing setting whereas $\mathcal{L}_\text{sky}$ is outward facing.

\section{High Dynamic Range}
\begin{figure*}[ht!]
    \centering
    \makebox[\textwidth]{%
        \begin{tikzpicture}
            \node (img) {\includegraphics[width=\textwidth]{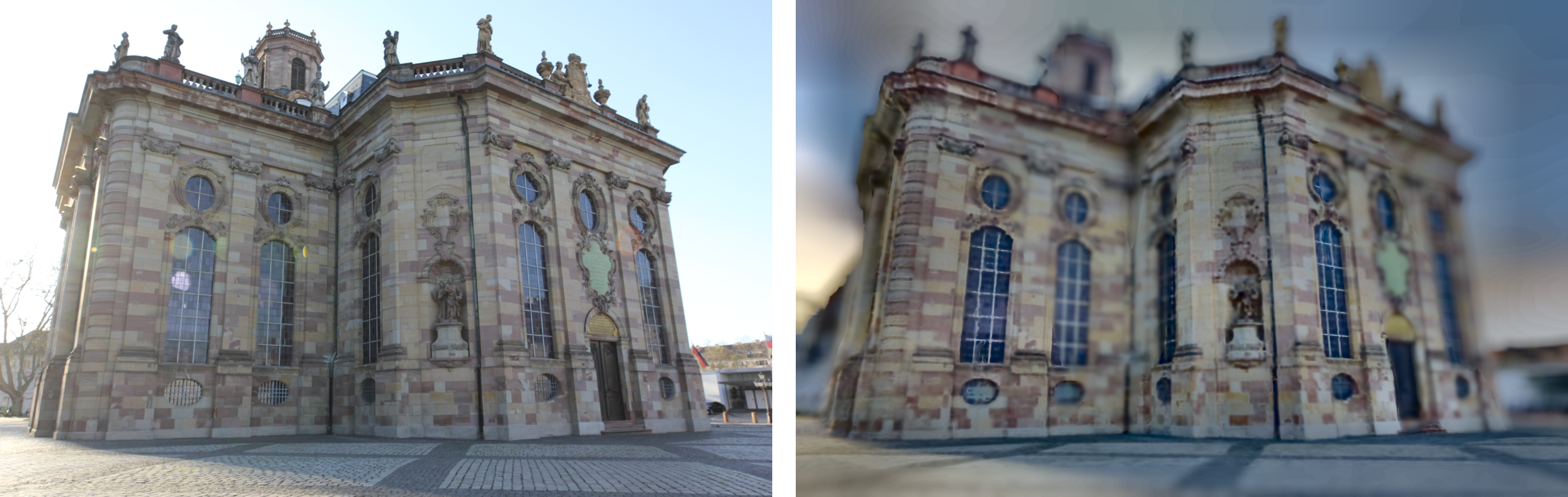}};
        \end{tikzpicture}
    }
    \caption{HDR post-processing capabilities of our model. LDR ground truth (left), depth-of-field and HDR tonemapping (right).}
    \label{fig:hdr_effects}
\end{figure*}

As our lighting and model are both optimised in linear HDR space we implicitly reconstruct HDR sky (and scene) from the LDR input images. This enables HDR post-processing of our renderings as shown in Fig \ref{fig:hdr_effects}.

\section{Additional Comparisons}
\begin{figure*}[ht!]
    \centering
    \makebox[\textwidth]{%
        \begin{tikzpicture}
            \node (img) {\includegraphics[width=\textwidth]{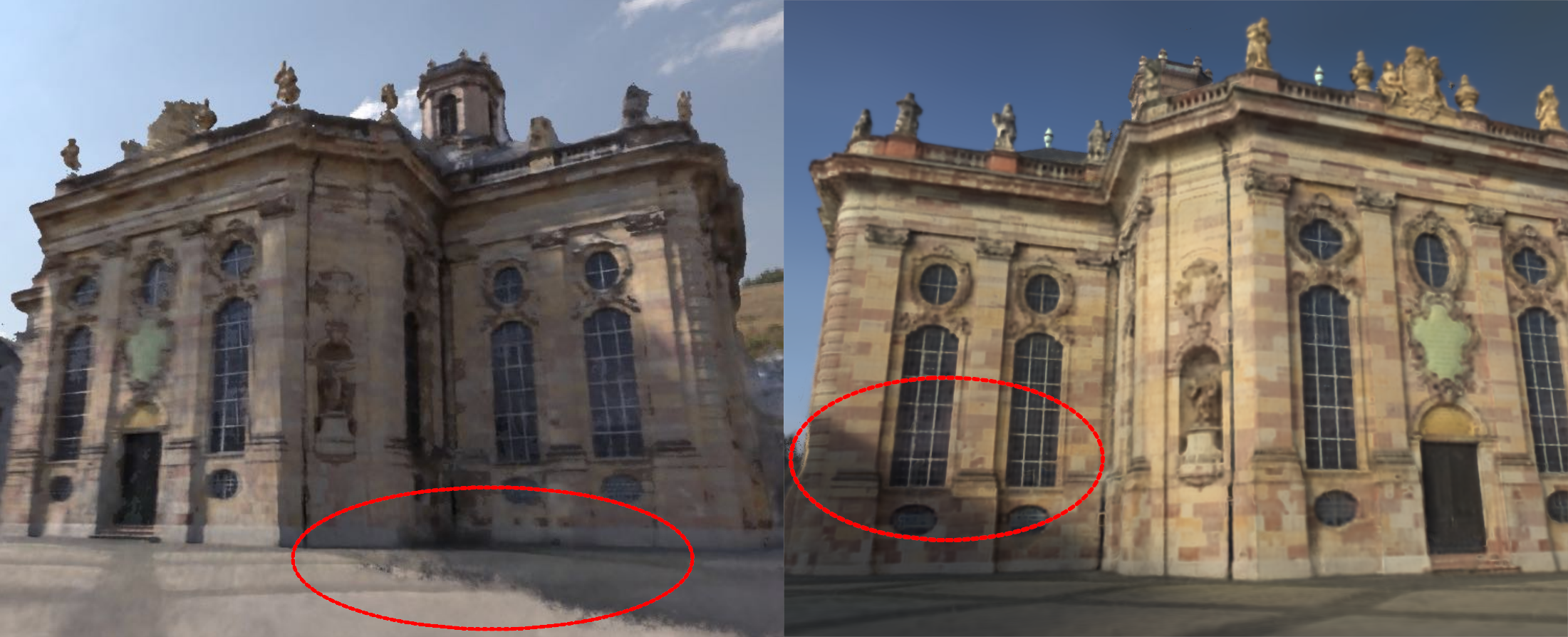}};
        \end{tikzpicture}
    }
    \caption{FERG (L) produces noisy shadows and is trained in a cascaded manner. Ours (R) produces sharp detailed shadows and is trained end-to-end.}
    \label{fig:fegr_poor_shadows}
\end{figure*}

In Figure \ref{fig:fegr_poor_shadows} we highlight the improvement in shadow quality our model provides over FEGR \cite{wang_neural_2023}. Our visibility and illumination models can represent high-quality sharp shadows whilst being trained concurrently with our geometry and albedo networks. Unlike FEGR which produces noisy artefacts and requires conversion of the SDF scene representation into a mesh for ray-tracing.

\label{sec:additional_comparisons}

\section{NeRF-OSR Relighting Benchmark}
\label{sec:relighting_eval_method}
The relighting benchmark for NeRF-OSR \cite{rudnev_nerf_2022}, in which the ground truth environment map from a session is used to relight the scene and the appearance error computed from a single viewpoint per session, for pixels within a provided mask, covers sites $1-3$. To benchmark NeuSky we chose to tackle a more challenging task and instead estimate our illumination environment from a single holdout image, a test image from another viewpoint of the scene during the same capture session. From the holdout viewpoint, we hold our model static and optimise only RENI++ latent codes and scale $\gamma$ for each holdout image. For this, we only optimise the appearance losses:
$$
\mathcal{L}_\text{eval\_illumination} = \mathcal{L}_\text{app} + \sum_{\mathbf{r}\in\mathcal{R}\cap\mathcal{S}_\text{sky}} \varepsilon(\mathbf{c}_\text{gt}(\mathbf{r}),\mathbf{c}_\text{sky}(\mathbf{r}))
$$
We then position the camera in the test viewpoint and evaluate within the provided mask. We decided to evaluate using this methodology for the following reasons. 
\begin{enumerate}
    \item The environment map to model alignment is unknown. SOL-NeRF \cite{sunSOLNeRFSunlightModeling2023} attempted to address this via rotations of the environment until the highest PSNR error was achieved and this was presumed to be the correct orientation.
    \item The images are not HDR, we discussed with the author of NeRF-OSR and their solution is an arbitrary scaling of saturated pixels, this scaling was set to $10$ for \cite{rudnev_nerf_2022} and $30$ in \cite{sunSOLNeRFSunlightModeling2023}, to simulate HDR before fitting their illumination model.
    \item Our method is more challenging as we estimate the illumination rather than being given provided with it. i.e.~we simultaneously evaluate illumination estimation and relighting.
\end{enumerate}
We, therefore, consider this the best tradeoff between accuracy and repeatability and recommend in the future others also use this evaluation method. We will make our fitted environment maps available for future evaluations.

\begin{figure*}[ht!]
    \centering
    \makebox[\textwidth]{%
        \begin{tikzpicture}
            \node (img) {\includegraphics[width=0.98\textwidth]{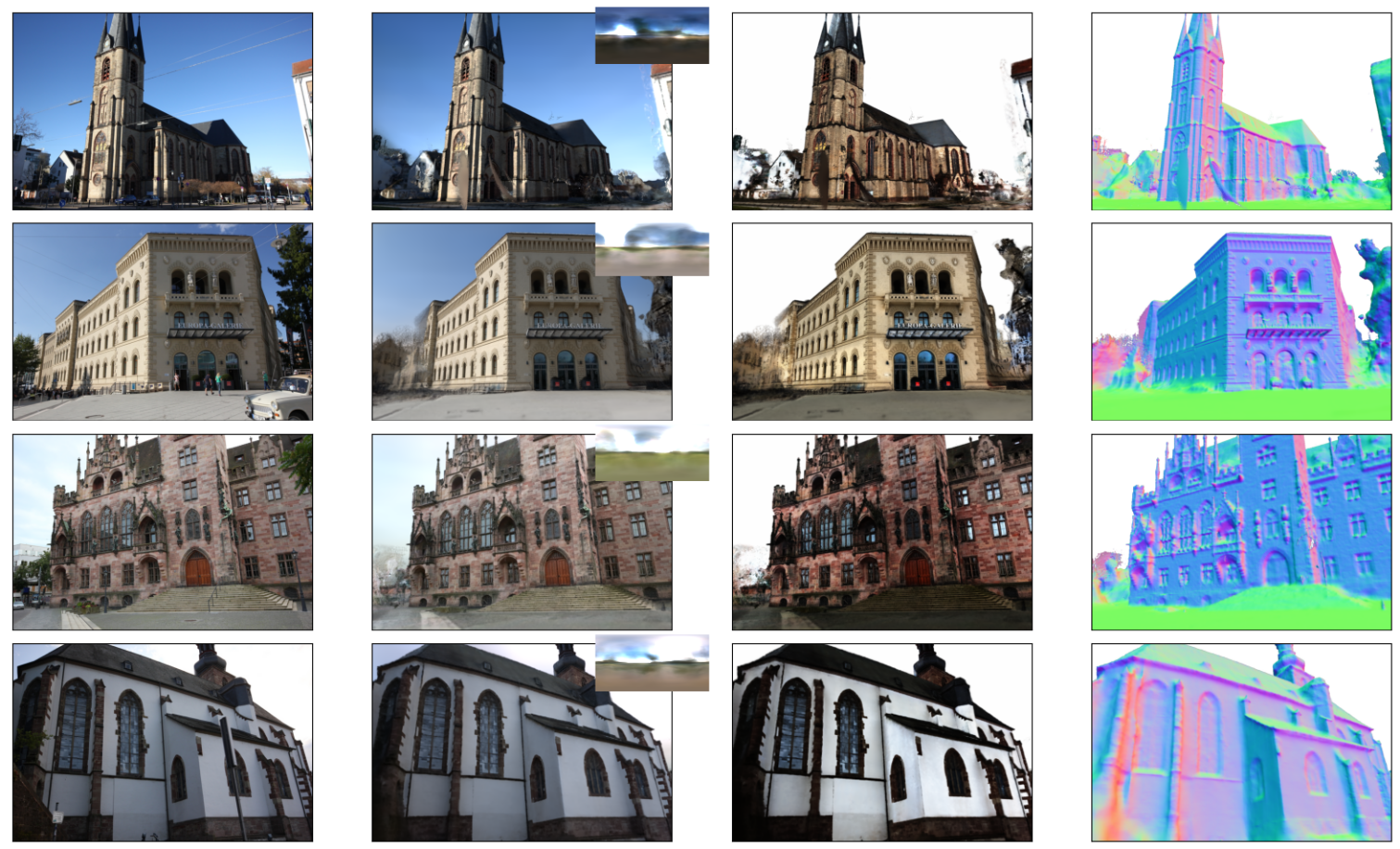}};
            \node at (-4.5, 3.8) {\small Ground Truth};
            \node at (-1.5, 3.8) {\small Render};
            \node at (1.5, 3.8) {\small Albedo};
            \node at (4.6, 3.8) {\small Normals};
        \end{tikzpicture}
    }
    \caption{Renders of four other scenes in NeRF-OSR \cite{rudnev_nerf_2022}. Estimated illumination of RENI++\cite{gardner_reni++_2023}, albedo and normals are shown alongside the ground truth images. Our method accurately disentangles albedo, lighting and shadows whilst producing very high-quality geometry.}
    \label{fig:more_examples}
\end{figure*}

\begin{figure*}[ht!]
    \centering
    \makebox[\textwidth]{%
        \begin{tikzpicture}
            \node (img) {\includegraphics[width=\textwidth]{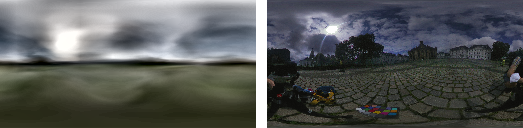}};
        \end{tikzpicture}
    }
    \caption{NeuSky lighting prediction (left), reference (right).}
    \label{fig:predicted_lighting}
\end{figure*}

\section{Implementation Details}

During data pre-processing, we assume all cameras are looking towards the object of interest and align the average focus point of all cameras to be at the centre of our scene.

As one of the classes in our CityScapes \cite{cordts_cityscapes_2016} segmentation masks is ground, we have an optional ground plane alignment loss that enforces consistency between the volume rendered normal and the world-up vector for rays inside that mask. We use the normal consistency loss from MonoSDF \cite{yuMonoSDFExploringMonocular2022}:
\begin{equation}
    \mathcal{L}_{\text{gp}}= \sum_{\mathbf{r}\in\mathcal{R}\cap\mathcal{S}_\text{gp}} \|N(\mathbf{r})-\mathbf{w}\|_1+\left\|1-N(\mathbf{r})^{\top} \mathbf{w}\right\|_1,
\end{equation}
where $\mathcal{S}_\text{gp}$ is the set of ground plane pixels, $N(\mathbf{r})$ is the volume rendered normal for ray $\mathbf{r}$ and $\mathbf{w}$ is the world-up vector defined as $[0, 0, 1]$. 

\section{Further Results and Videos}

In Figure \ref{fig:more_examples} we provide more renderings of our model fit to the remaining NeRF-OSR \cite{rudnev_nerf_2022} scenes, demonstrating further our model's ability to capture high-frequency geometric details, and accurately disentangle shading and albedo ambiguities. We encourage the reader to view the rendered videos on our project page demonstrating the multi-view and re-lighting capabilities of our model. In each video, we move the camera around the scene, once the camera comes to a stop we rotate the illumination environment to demonstrate the accurate shadow reproduction and relighting capabilities of our model. We also include videos of our Directional Distance Field (DDF) which we used for our sky visibility estimations and is trained concurrently with the scene representation. Each frame of this video is a single forward pass through our DDF which is able to produce the highly accurate depth maps required for accurate shadows.

%
%
\bibliographystyle{splncs04}
\bibliography{references}